\documentclass[10pt,journal,compsoc]{IEEEtran}

\usepackage{times}
\usepackage{epsfig}
\usepackage{graphicx}
\usepackage{amsmath}
\usepackage{amssymb}
\usepackage{algorithm}
\usepackage{algorithmic}
\usepackage{footnote}
\usepackage{epstopdf}
\usepackage{subfigure}
\usepackage{hyperref}
\usepackage{multirow}
\usepackage{hhline}

\newcommand{\ie}{\textit{i}.\textit{e}.}
\newcommand{\eg}{e.g.}
\newcommand{\norm}[1]{\left\lVert#1\right\rVert}

\ifCLASSOPTIONcompsoc
  \usepackage[nocompress]{cite}
\else
  \usepackage{cite}
\fi
\ifCLASSINFOpdf
\else
\fi
\begin{document}
%
\title{An Experimental Evaluation of Covariates Effects on Unconstrained Face Verification}
\author{Boyu~Lu,~\IEEEmembership{Student Member,~IEEE}, Jun-Cheng~Chen,~\IEEEmembership{Member,~IEEE}, Carlos~D~Castillo,~\IEEEmembership{Member,~IEEE} and~Rama~Chellappa,~\IEEEmembership{Fellow,~IEEE}
\thanks{B. Lu is with the Department of Electrical and Computer Engineering, University of Maryland, College Park, 20742, USA (e-mail: bylu@umiacs.umd.edu)}

\thanks{J. Chen is with the Center for Automation Research, University of Maryland, College Park, 20742, USA (e-mail: pullpull@cs.umd.edu)}

\thanks{C. D. Castillio is with the Center for Automation Research, University of Maryland, College Park, 20742, USA (e-mail: carlos@umiacs.umd.edu)}

\thanks{R. Chellappa is with the Department of Electrical and Computer Engineering and the Center for Automation Research, University of Maryland, College Park, 20742, USA (e-mail: rama@umiacs.umd.edu)}
}

%
%

\markboth{Journal of \LaTeX\ Class Files,~Vol.~14, No.~8, August~2015}%
{Lu \MakeLowercase{\textit{et al.}}: An Experimental Evaluation of Covariates Effects on Unconstrained Face Verification}
%



\IEEEtitleabstractindextext{%
\begin{abstract}
Covariates are factors that have a debilitating influence on face verification performance. In this paper, we comprehensively study two covariate related problems for unconstrained face verification: first, how covariates affect the performance of deep neural networks on the large-scale unconstrained face verification problem; second, how to utilize covariates to improve verification performance. To study the first problem, we implement five state-of-the-art deep convolutional networks (DCNNs) for face verification and evaluate them on three challenging covariates datasets. In total, seven covariates are considered: pose (yaw and roll), age, facial hair, gender, indoor/outdoor, occlusion (nose and mouth visibility, eyes visibility, and forehead visibility), and skin tone. These covariates cover both intrinsic subject-specific characteristics and extrinsic factors of faces. Some of the results confirm and extend the findings of previous studies, others are new findings that were rarely mentioned previously or did not show consistent trends. For the second problem, we demonstrate that with the assistance of gender information, the quality of a pre-curated noisy large-scale face dataset for face recognition can be further improved. After retraining the face recognition model using the curated data, performance improvement is observed at low False Acceptance Rates (FARs) (FAR=$10^{-5}$, $10^{-6}$, $10^{-7}$).
\end{abstract}

}

\maketitle

\IEEEdisplaynontitleabstractindextext

%
\IEEEpeerreviewmaketitle

\section{Introduction}
%
%
%
%


\IEEEPARstart{F}{ace} Verification has been receiving consistent attention in computer vision community for over two decades~\cite{zhao2003face}. The task of face verification is to verify whether a given pair of face images/templates belongs to the same subject. Recently, due to the rapid development of deep convolutional neural networks (DCNNs), face verification performance has surpassed human performance in most controlled situations and some unconstrained cases~\cite{taigman2014deepface,sun2014deep,schroff2015facenet,chen2015end}. Although deep features have proven to be more robust to moderate variations in pose, aging, occlusion and other factors than hand-crafted features, some recent works have noticed that face verification performance is still significantly affected by many covariates~\cite{parde2017face,sun2015deeply,du2015cross,ranjan2017l2}. 

Covariates are factors that usually have an undesirable influence on face verification performance (\eg, gender induces different human facial appearance characteristics in nature.). Some covariates represent different aspects of faces such as pose, expression and age, while some other covariates represent subject-specific intrinsic characteristics like gender, race and skin tone, and other covariates reflect extrinsic factors in images, such as illuminations, occlusion and resolution. Analyzing the effects of these covariates can not only help understand fundamental problems in face verification, but also provide insights to improve existing face verification algorithms. 

Previous studies have analyzed many covariates effects on face recognition performance~\cite{lui2009meta,abdurrahim2017review,paone2011difficult,beveridge2009factors}. However, most of them are outdated, and there are several reasons why a new study on these covariates is needed. First, most studies have been conducted before the emergence of deep networks. Since deep networks have significantly improved the robustness of features against many covariates, it is unclear whether the results of covariate effects concluded from hand-crafted features are still valid when deep features are used. Second, most datasets studied in previous works are small (\eg, 41,368 images from 68 people in CMU PIE~\cite{sim2002cmu} dataset.) and the class distribution of some covariates is severely imbalanced. In this situation, some conclusions may become statistically biased. Moreover, due to the absence of large data, very few experiments have studied covariate effects at extremely low FARs ($10^{-5}$, $10^{-6}$). Third, the face images in former studies were captured in a constrained environment (\eg, CMU PIE dataset~\cite{sim2002cmu}), which is less applicable in practice. Last but not least, most existing papers only focus on whether some covariate values have advantages over other values (\eg, whether a male is easier to recognize than a female), but few of them try to exploit covariate information to improve face verification performance. In fact, some covariates (\eg, gender, race) contain subject-specific information of faces, and are more robust to many extrinsic variations than low-level features. Properly exploiting them could measurably improve face verification performance significantly~\cite{hu2017attribute}.

In this paper, we investigate two important problems: a) how different covariates affect the performance of state-of-the-art DCNNs for unconstrained face verification; b) how to utilize covariate information to improve face verification performance. For the first problem, we implement five state-of-the-art face DCNNs and evaluate them on three challenging covariates protocols: 1:1 covariates protocol of the IARPA JANUS Benchmark B (IJB-B) dataset~\cite{ijbb} and its extended version the IARPA JANUS Benchmark C (IJB-C)~\cite{ijbc}, and Celebrity Frontal-Profile Face datasets~\cite{sengupta2016frontal}. We report the performance of each individual network and the performance of the score-level fusion method. We also compare the results with some other well-known and publicly available deep face networks. Among the datasets, IJB-C 1:1 covariate protocol is currently the largest public covariate dataset for unconstrained face verification. The protocol contains seven covariates covering different covariate types. Moreover, the IJB-C dataset is designed to have a more uniform geographic distribution of subjects across the globe, which makes it possible to carefully evaluate many covariates (\eg, like age and skin tone) in details. 

By conducting extensive experiments on IJB-B and IJB-C datasets, we observe many interesting results for different covariates. Some of our findings support conclusions drawn from previous studies. For example, extreme yaw angles do substantially degrade the performance~\cite{sengupta2016frontal} and outdoor images are harder to be recognized~\cite{liu2007outdoor}. Meanwhile, we also find some results which extend the findings of previous works due to the availability of larger datasets. For instance, most previous studies show that face recognition algorithms usually achieve better performance on elder subjects than younger subjects~\cite{lui2009meta,beveridge2009factors}. But in their studies, most of the enrolled subjects are under 40 years old. However, our experiments with much more subjects with a wider age range show that the performance does not monotonically increase as age progresses. The performance increases from age group $[0,19]$ to age group $[35,49]$ but begins to drop for age group $[50,65]$ and $65+$. The results demonstrate that neither too young nor too old people are easy to recognize, but the recognition results for very young people (\ie, $[0,19]$) are worse. Moreover, we are able to better evaluate some covariates like gender where previous works came to contradictory conclusions~\cite{lui2009meta}. Our experiments show that males are easier to be verified than females in general. However, when we combine gender with other covariates (age, skin tone) to investigate their mixed effects, we find that the face verification performance for females becomes better than males' for older age group and darker skin tones. Finally, some of our results are surprising yet rarely analyzed in other papers. One example is that roll variations greatly affect verification performance in unconstrained situation. Since most previous studies may have used manually aligned faces, roll variation was not a significant factor in their studies. However, in unconstrained environments, face alignment becomes a key component and our finding shows that the performance variations might result from the fact that face alignment algorithms fail to work perfectly for faces in extreme roll angles.

For the second problem, we utilize gender information to curate a noisy large-scale face dataset. Specifically, we find that the curated MS-Celeb1M~\cite{guo2016msceleb,lin2017proximity} still contains many noisy labels where some subjects are still mixed with images from different genders. Training using data with these noisy labels may potentially hurt the discriminative capability of deep models and degrade their performance, especially in low FAR regions (10-5, 10-6, etc). Therefore, we leverage gender information to further curate the training set and remove those subjects mixed with images of both males and females. First, we predict gender probability for each image in the training set using the multi-task face network proposed in~\cite{ranjan2017all}. Since gender prediction may become inaccurate when gender probabilities are near 0.5, we only consider faces with gender probability greater than 0.6 (male) or smaller than 0.4 (female). Then for each subject, if the percentage of faces from the minority gender exceed the threshold (3\% in our experiment), we remove all the face images of the subject from the training set. After retraining the model using the curated data, the performance improves at low FARs.

The main contributions of this paper are summarized as follows:

\begin{itemize}
  \item We comprehensively study the effects of seven covariates on the performance for unconstrained face verification. The datasets we use are the largest public covariate face datasets, which allows evaluation at very low FARs ($10^{-5}$, $10^{-6}$). We test all the covariates using the state-of-the-art deep models. This gives an insight on the limitations of many existing deep CNNs for face covariates.

  \item We study the mixed effects of multiple covariates. This is an important problem for unconstrained environment yet not deeply explored by previous studies.

  \item We propose to utilize gender information to help effectively curate the training data and achieve better performance. 

\end{itemize}

The rest of the paper is organized as follows. A brief review of literatures on covariate analysis for face verification is presented in Section~\ref{related}. In Section~\ref{system}, we introduce five state-of-the-art DCNNs for face verification and the ways to fuse the similarity scores from them. A method for how to utilize gender covariate for training set curation is presented in Section~\ref{clean_gender}. Experimental results on three different covariates datasets are shown in Section~\ref{experiments} and finally we summarize and conclude the paper in Section~\ref{conclude}.

\begin{figure*}
    \begin{center}
            \includegraphics[width=0.95\textwidth]{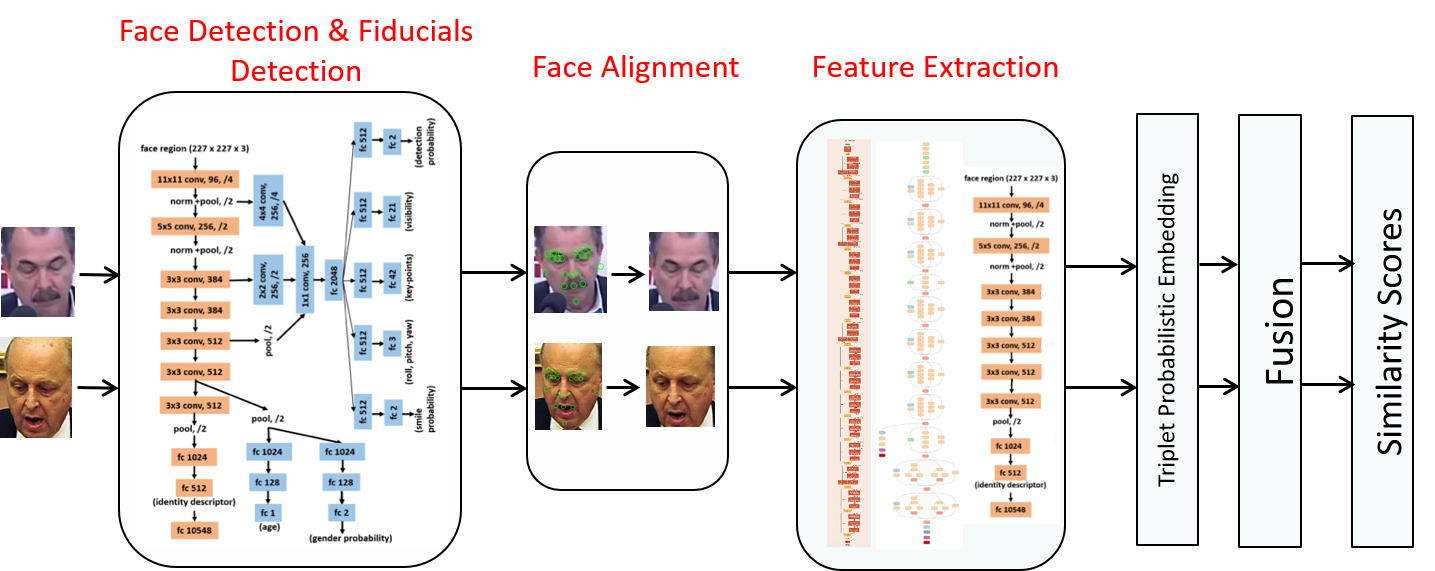}
    \caption{System pipeline for unconstrained face verification.}
    \label{fig:system}
    \end{center}  
\end{figure*}

\section{Related Works}
\label{related}
Several prior works discussed the effects of covariates on face recognition performance~\cite{gross2001quo,beveridge2009factors,beveridge2010frvt,lui2009meta,givens2013introduction,abdurrahim2017review,grm2017strengths}. Gross~\emph{et al.}~\cite{gross2001quo} evaluated two algorithms on three face datasets and discussed five covariates: pose, illumination, expression, occlusion and gender. They varied each covariate with other factors fixed and examined performance changes for two algorithms. Similarly, Beveridge~\emph{et al.}~\cite{beveridge2009factors,beveridge2010frvt} applied a statistical approach called the Generalized Linear Mixed Model (GLMM) to analyze two types of covariates: subject covariates (\eg, gender, race, wearing glasses) and image covariates (\eg, image size ratio, the number of pixels between eyes). Three algorithms were tested, and they claimed that effects of covariates for different algorithms varied significantly. In~\cite{givens2013introduction}, Givens~\emph{et al.} split faces into three groups (good, bad and ugly) based on the performance of their verification rates. They used GLMM to analyze the underlying effects of different covariates over these three groups. They showed that many covariate effects on verification performance are universal across three groups. Different from the previous works that use statistical methods to analyze the covariates, Lui~\emph{et al.}~\cite{lui2009meta} presented a meta-analysis for six covariates on face recognition performance by summarizing and comparing different papers. In order to guarantee that the conclusions are meaningful, they restricted their analysis to frontal, still, visible light images. In~\cite{abdurrahim2017review}, Abdurrahim \emph{et al.} reviewed recent research on demographics related covariates (age, race, and gender). They drew similar conclusions as in~\cite{lui2009meta} for most covariates (\eg, age, gender) while they also pay attention to interactions among demographics covariates. In~\cite{grm2017strengths}, Grm \emph{et al.} analyzed the effects of some covariates related to image quality (like blur, occlusion, brightness) and model characteristics (like color information). They used the Labeled Face in the Wild (LFW)~\cite{LFWTech} dataset to synthesize degraded images and compare the robustness of four widely used DCNNs to each covariate. In the following subsections, we briefly review the main findings of related works for each specific covariate.

\subsection{Pose}

Studies on effects of pose variations on face recognition have been reported in~\cite{ding2016comprehensive,gross2010multi,li2014maximal,sun2014deep}. Pose variations generally involve yaw, roll and pitch. Among them, yaw and pitch variations are out-of-plane rotations while roll corresponds to in-plane rotation. Normally, roll variations can be eliminated by applying face alignment using similarity or affine transform to warp the face into pre-defined canonical coordinates while yaw and pitch variations are much harder to rectify and thus have a larger impact on the performance than roll in face recognition. Recent studies show that even the best deep-learning based face models are still severely affected by large pose variations~\cite{li2014maximal,sun2014deep}.

\subsection{Age}
The effects of age on face verification performance are usually studied in two ways: aging and age groups. Aging effects are best analyzed in cross-age face verification scenario because it tries to recognize faces from different ages for the same subject. This is a challenging problem because for most subjects their face appearance changes tremendously as they become older~\cite{ramanathan2006face,ling2007study,best2018longitudinal}. In contrast, age group effects refer to the difficulty in recognizing people from different age groups. This study aims to explore whether a certain age group is harder to be recognized than other groups~\cite{lui2009meta,beveridge2010frvt,givens2013introduction,best2018longitudinal}. 

\subsubsection{Aging}

It has been revealed by almost all studies that age variations for the same person impair verification performance. However, the effects may not be significant if the age differences are within several months~\cite{guo2010cross}. Although aging effects become substantial if the acquisition time difference exceeds several years, there are still some features preserved on faces that can be utilized for face verification~\cite{best2018longitudinal}. Some works tried to reduce the intra-subject aging effects for face verification by discriminative learning or feature selections~\cite{ramanathan2006face,du2015cross}. Best-Rowden \emph{et al.} applied the mixed-effects models to analyze aging effects using a large mugshot dataset. They showed that the average similarity score of genuine pairs decreases significantly with increasing elapsed time between a gallery and probe. However, they found that on average the genuine pairs can still be recognized at FAR = 0.01\%, when the elapsed time is no more than 15 years.

\subsubsection{Age groups}

The effects of age groups have been discussed by many studies. Interestingly, different from many other covariates where different studies show different results, most studies have come to similar conclusions on age group effects: older subjects are usually easier to recognize than younger subjects~\cite{beveridge2009factors,beveridge2010frvt,lui2009meta,abdurrahim2017review}. However, most of the experiments were conducted in an environment where age distributions are very imbalanced and the number of samples for young people is much larger than old people. The imbalance increases the difficulty of verification for young people. In~\cite{ho2007younger}, Ho~\emph{et al.} did experiments with each age group evenly distributed. They found that the performance for young ages and old ages did not show statistically significant difference.

\subsection{gender}
Gender is one of the intrinsic characteristics of a human face. A man's face is different from a woman's face in terms of shape, facial part distance and facial hair. However, studies on the effects of gender on verification performance have led to different conclusions. Lui~\emph{et al.}~\cite{lui2009meta} summarized covariates research papers from 2001 to 2008. Seven studies found men were easier to recognize~\cite{beveridge2009factors,beveridge2010frvt}, while five claimed women were easier~\cite{beveridge2009factors,beveridge2010frvt,beveridge2008focus}, and six reported that gender shows no effects on face recognition performance~\cite{beveridge2009factors,beveridge2010frvt,givens2004features,givens2003statistical}. More recently, Grother~\emph{et al.}~\cite{grother2010report} evaluated seven commercial algorithms and five of them were more accurate on males. On the other hand, gender is also shown to correlate with other covariates like age~\cite{lui2009meta}. In~\cite{phillips2003evaluation}, Phillips~\emph{et al.} reported that performance difference between males and females decreases as people age.

\subsection{Race and skin tone}

Race and skin tone are also demographic covariates that represent subject-specific characteristics of people. There were several studies on the effects of races and skin tones on face verification performance, but few of them can be clearly interpreted~\cite{lui2009meta,abdurrahim2017review}. This is mainly due to the fact that most datasets are very biased with respect to race distribution. In~\cite{lui2009meta}, all the datasets they studied contain more Caucasians than East Asians with a ratio of 3 to 1. Therefore, even if East Asian outperforms Caucasians in all the cases in~\cite{lui2009meta}, it is still hard to conclude that East Asians are easier for verification. In another paper~\cite{grother2010report}, Grother~\emph{et al.} found that the influences of race on the performance are conflicted for different algorithms. African American are more easily recognized than Caucasians for five out of six algorithms. American Indians and Asian are easier to be recognized for three algorithms but are more difficult for one algorithm. These results may be simply due to different training processes where algorithms are superior for some races over others. There is also one paper studying the influence of skin tones on face verification~\cite{bar2009role}. Bar-Haim~\emph{et al.} reported that the effects of skin tone on verification performance are not as important as other unique facial features for certain races.

\subsection{Occlusion}

Occlusion could be caused by wearing glasses/sunglasses, masks, scarves or by hairstyle (like bangs). It has been widely investigated that occlusion of key facial parts can substantially degrade the verification performance~\cite{beveridge2009factors,beveridge2010frvt,sun2015deeply,grm2017strengths}. However, different algorithms are not sensitive to occlusions to the same degree~\cite{beveridge2010frvt,sun2015deeply,grm2017strengths}. There is also one study reporting that consistently wearing glasses may help improve verification performance for faces acquired in outdoors~\cite{beveridge2010frvt}. 

\subsection{Indoor/Outdoor}

The effects of indoor/outdoor are related to some other image covariates like illumination, resolution, and blur. Most studies revealed that indoor performance is generally better than outdoor~\cite{beveridge2008focus,beveridge2010frvt,lui2009meta,liu2007outdoor}. Moreover, the indoor/outdoor effect is also found to correlate with other covariates. In~\cite{beveridge2008focus,beveridge2010frvt}, Beveridge~\emph{et al.} reported that recognition performance under outdoor environments often favors high resolution images while for low resolution images, indoor environments are preferred. Another finding reported in~\cite{beveridge2010frvt} is that indoor/outdoor taxonomy also affects verification performance for different genders and sometimes may even reverse the trends.

\subsection{Facial Hair}

Studies on facial hair effects are limited compared to other common covariates. Earlier studies~\cite{givens2003statistical,givens2004features} suggested that performance is better when facial hair exists in at least one of the images. However, the underlying reason for this result is unclear because facial hair is not a unique biometric for recognition and can be changed easily. 

\section{Evaluation Pipeline Overview}
\label{system}
Before addressing the first problem, we briefly introduce the five deep networks that we used to perform unconstrained face verification over covariates. Before feeding a face image into these networks, preprocessing steps including face detection, facial landmark detection and face alignment are performed by using the multi-task CNN framework proposed in~\cite{ranjan2017all}. More details about the multi-task CNN are provided in Section~\ref{sec:all_in_one}. After feature extraction, we applied triplet probabilistic embedding (TPE)~\cite{sankaranarayanan2016triplet} on the deep features to further improve the face verification performance. The TPE aims to learn a projection matrix $\mathbf{W}$ by minimizing a negative log-likelihood objective function. The idea of TPE is to push positive pairs closer and negative pairs farther apart by selecting anchor and positive/negative samples. More details can be found in~\cite{sankaranarayanan2016triplet}. The end-to-end system pipeline is illustrated in Figure~\ref{fig:system}.



\subsection{Deep Representations for Faces}

To capture the different characteristics of faces, we use features extracted from five state-of-the-art deep neural networks. These five networks have different architectures and training sets with their own strengths and weaknesses. Details of each network architecture are presented next.

\subsubsection{Training set preparation}
\label{sec:curation}
To train the deep networks, we use UMD-Faces~\cite{bansal2016umdfaces,bansal2017s}, Megaface~\cite{nech2017level}, and MS-Celeb-1M~\cite{guo2016msceleb}. In addition, we found that directly using the original MS-Celeb-1M dataset for training does not achieve good performance because the labels are very noisy. Therefore, we used a curated version of MS-Celeb-1M dataset which is done using a clustering method introduced in~\cite{lin2017proximity}. The curated dataset contains about 3.7 millions face images from 57,440 identities. After curation, many noisy labels are removed while sufficient amount of face images with different variations are retained.

\subsubsection{CNN-1}
This network employs ResNet-27 model introduced in~\cite{wen2016discriminative}. We modify the original model by removing the center loss and replacing the softmax loss with the $L_{2}$-softmax loss introduced in~\cite{ranjan2017l2}. The $L_{2}$-softmax loss enables the learned features to lie on a hypersphere with a fixed radius $\alpha$ before feeding into the softmax classifier. Since the norm of the features for hard samples is usually smaller than easy samples when applying softmax loss~\cite{ranjan2017l2}, enforcing the $L_{2}$-softmax loss ensures that the training process will focus more on hard samples and significantly improves the verification performance~\cite{ranjan2017l2}. In addition, we also add one more fully connected layer with 512-D before $L_{2}$-softmax layer to reduce the feature dimension and the total number of model parameters. For the input size, we change the original size of $112\times96$ to $128\times128$ for improved face alignment. To train the model, we use a curated version of the MS-Celeb-1M dataset described in Section~\ref{sec:curation}, which contains $3.7$ million images from $57,440$ subjects.

\subsubsection{CNN-2}

The second network uses the ResNet-101~\cite{he2016deep} architecture as the base network. CNN-2 is deeper than CNN-1 and has larger input size of dimensions $224\times224$. The basic blocks for CNN-2 use bottleneck structures to reduce the number of model parameters and achieve deeper networks given certain memory constraints. Similar to CNN-1, CNN-2 also replaces the original softmax loss with the $L_{2}$-softmax loss and adds an extra fully connected layer before the $L_{2}$-softmax layer. CNN-2 is trained using two different training sets and thus two different models are obtained. One model is called CNN-2\_S because a small training set is used (curated MS-Celeb-1M dataset) and the other model is called CNN-2\_L because it uses a larger training set (curated MS-Celeb-1M dataset, and Megaface).

\begin{figure}[tb]
\begin{center}
 \includegraphics[width=3.5in]{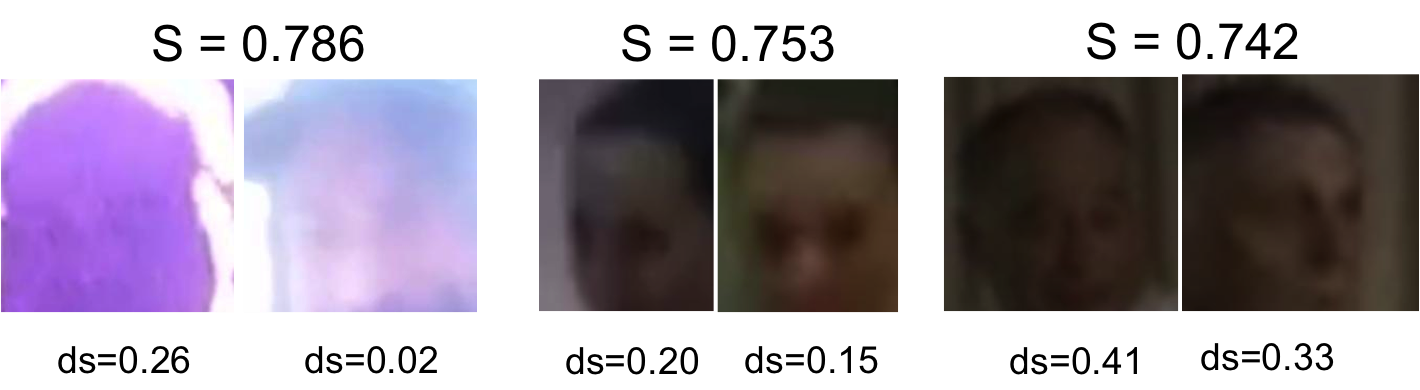}
 \caption{Examples of hard negative pairs for low detection confidence but have high similarity scores. $ds$ indicate the detection scores for the images and $S$ represents similarity score for each pair.}
 \label{fig:low_ds}
\end{center}
\end{figure}

\subsubsection{CNN-3}

The Inception-ResNet-v2~\cite{szegedy2017inception} model is used as the base network. This model combines the inception architecture with residual connections and achieved state-of-the-art performance on the ImageNet classification challenge. In addition, scaling layers are also included in the network architecture which scale down the residuals for more stable training. We adapt the Inception-ResNet-v2 model by adding a 512-D fully connected layer before the last layer. The training set contains over six millions images from about 58,000 subjects. These images includes a mixture of about 3.7 million still images from the curated MS-Celeb-1M dataset in Section~\ref{sec:curation}, about 300,000 still images from the UMDFaces dataset~\cite{bansal2016umdfaces}, and about 1.8 million video frames from the extension to the UMD-Faces Video dataset~\cite{bansal2017s}.  


\subsubsection{CNN-4}
\label{sec:all_in_one}
This network is based on the all-in-one CNN architecture~\cite{ranjan2017all}. The model is trained in a multi-task learning framework which utilizes the correlations among different tasks to learn a more robust model than learning each task individually. Lower layers of the network are shared for all the tasks to produce a generic representation while intermediate layers are only shared among more related tasks. Each task also has its task-specific layers and losses. In this paper, we mainly utilize the face detection, facial landmark detection branches for face alignment, and the face recognition branch to generate face features. We also use the gender classification branch to estimate gender probabilities. The face detection and facial landmark detection branches share the first six layers and have two separate fully connected layers for each task. The face recognition branch consists of seven convolutional layers followed by three fully connected layers. Same training set is used as for CNN-1 and CNN-2\_S.

\subsection{Face Matching and Score Level Fusion}

After we obtained the extracted features from the learned deep networks and the embedding matrix $\mathbf{W}$ from TPE~\cite{sankaranarayanan2016triplet}, the similarity scores for each pair $\{x_i, x_j\}$ is computed by simply using the cosine similarity of the two embedded features:

\begin{equation}
\label{cos_sim}
s_{ij} = \frac{(\mathbf{W}x_i)^T(\mathbf{W}x_j)}{\norm{\mathbf{W}x_i}\norm{\mathbf{W}x_j}}
\end{equation}

In the last stage of the proposed system, we fuse the scores computed from the five networks as the final similarity score. We observe that the similarity scores may become unreliable when the image qualities are poor. Meanwhile, we find the face detection scores obtained from the face detection branches of the CNN-4 is a good indication of image quality. More specifically, low detection scores usually indicate low quality of the detected faces (low resolution, extreme pose or severe blur) and the deep models failed to extract useful facial features from theses faces. Figure~\ref{fig:low_ds} shows some hard negative pairs with low detection scores but high similarity scores. We notice that the main reason for the high similarity scores is that these pairs are all very blurred and each pair has similar background. To address this issue, we reweight the similarity scores when the face detection scores of the corresponding pairs are low.


\begin{equation}
\label{score_reweight}
\hat{s_i}=\left\{ \begin{array}{ll}
s_i\mbox{,} &\mbox{if } ds> thr\\
\alpha s_i\mbox{,} &\mbox{otherwise,}
\end{array}\right. \\
\end{equation}

where $ds$ is the minimum of the detection scores for the pair of faces, $thr$ is the threshold, $\alpha$ is the reweight coefficient. 

Then we simply average the reweighted similarity scores from the five networks to get the final results.

\begin{equation}
\label{score_fuse}
s = \frac{1}{5}\sum_i^5\hat{s}_i
\end{equation}


%

\begin{figure}[tb]
\begin{center}
 \includegraphics[width=3.2in]{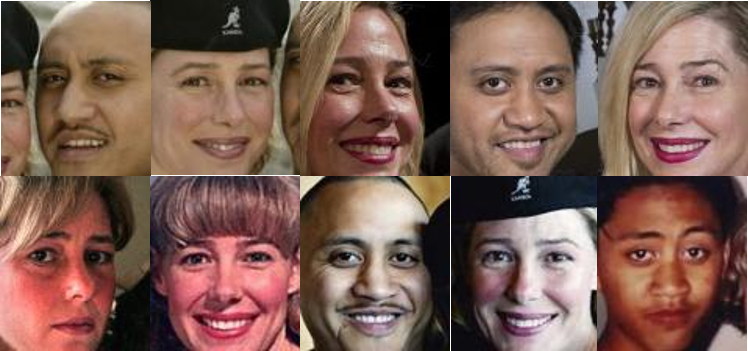}
 \caption{Instances of subjects with noisy labels}
 \label{fig:noisy}
\end{center}
\end{figure}

\section{Performance Improvement by Exploiting Gender Information}
\label{clean_gender}

Although many noisy labels are removed after curating the training set using the clustering method as mentioned in Section~\ref{sec:curation}, there still exists many noisy labels which cannot be handled by clustering. Figure~\ref{fig:noisy} shows an example of a subject which still contains noisy labels. We can see that those mislabeled face images look very similar to the correctly-labeled faces. Since the clustering method mainly determines a cluster based on the appearance similarity between faces, it is very hard to discover these mislabeled images. 

However, we observe that some mislabeled images have different genders compared to the correctly-labeled images. This motivates us to further curate the training set by exploiting the gender information. First, gender probabilities are estimated using the all-in-one face network~\cite{ranjan2017all} for all the face images in the pre-curated MS-Celeb-1M dataset in~\ref{sec:curation}. Since gender estimation may become unreliable when gender probabilities are near 0.5, we only consider faces with gender probability greater than 0.6 (male) or smaller than 0.4 (female). For each subject, if the number of faces from the minority gender accounts for more than 3\% of the total number of faces, we eliminate the whole subject. In total, we removed 248,059 faces from 4,160 subjects.

\begin{figure}[tb]
\begin{center}
 \includegraphics[width=3.5in]{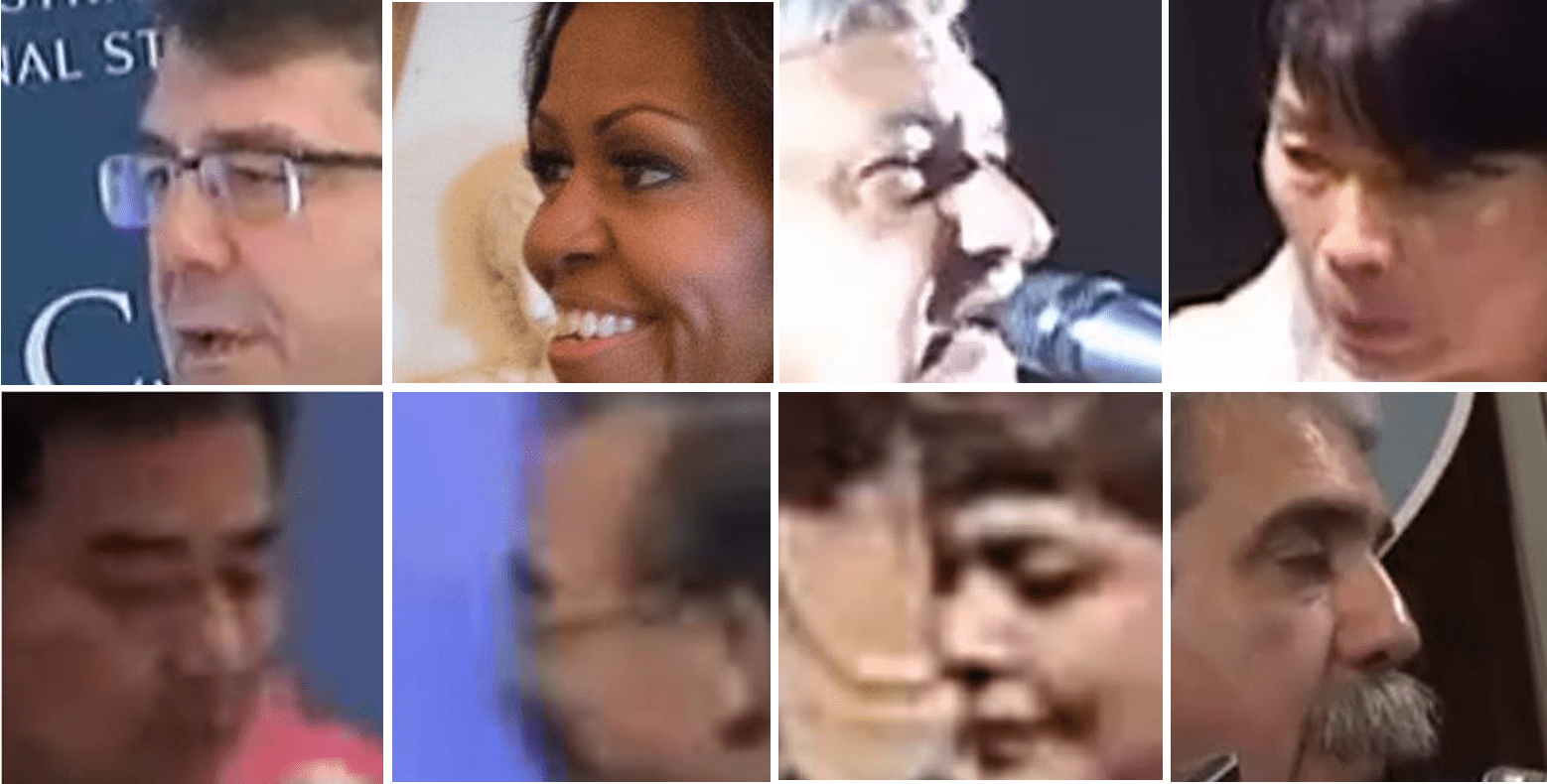}
 \caption{Sample images for IJB-B and IJB-C datasets.}
 \label{fig:sample_ijb}
\end{center}
\end{figure}

\section{Experimental Results}
\label{experiments}

To analyze the covariate effects on unconstrained face verification performance, we evaluated the five deep networks on three challenging face datasets that have face verification covariate protocols: the IARPA JANUS Benchmark B (IJB-B) 1:1 covariates~\cite{ijbb}, the IARPA JANUS Benchmark C (IJB-C) 1:1 covariates~\cite{ijbc} and the Celebrities in Frontal-Profile in the Wild (CFP)~\cite{sengupta2016frontal}. IJB-B and IJB-C 1:1 covariates both contain seven covariates protocols while the CFP dataset mainly focuses on extreme pose variations.

\subsection{IJB-B and IJB-C 1:1 covariates protocol}

The IARPA JANUS Benchmark B (IJB-B) dataset~\cite{ijbb} is a moderate-scale unconstrained face dataset with face detection, recognition and clustering protocols. It consists of 1845 subjects with human-labeled ground truth face bounding boxes, eye/nose locations, and covariate meta-data such as occlusion, facial hair, and skin tone for 21,798 still images and 55,026 frames from 7,011 videos. The 1:1 covariates protocol of IJB-B aims to analyze the effects of seven different covariates (i.e., pose (yaw and roll), age, facial hair, gender, indoor/outdoor, occlusion (nose and mouth visibility, eyes visibility, and forehead visibility), and skin tone.) on face verification performance. The protocol has 20,270,277 pairs of templates (3,867,417 positive and 16,402,860 negative pairs) which enables us to evaluate algorithms at low FARs region of ROC curve (\eg, FAR at 0.001\% and 0.0001\%). Each template contains only one image or a video frame. Some sample images are shown in the first row of Figure~\ref{fig:sample_ijb}. The IARPA JANUS Benchmark C (IJB-C) dataset~\cite{ijbc} is an extended version of the IJB-B dataset, and contains more subjects and pairs for evaluation. It consists of 3,531 subjects containing 140,739 images and video frames. The 1:1 covariates protocol has 47,404,001 pair of templates (7,819,362 positive and 39,584,639 negative pairs). Some sample images are shown in the second row of Figure~\ref{fig:sample_ijb}. 

To understand the effects of different covariates on face verification performance, in addition to the identity label (positive or negative) for each pair of templates, covariate labels are also assigned to each pair. To analyze a certain covariate (like gender), all pairs are split into groups based on the value of covariate labels (female = 0 and male = 1). The ROC curve is drawn for each group and the performance difference among different groups reflects the effects of the covariates. When we evaluate the general performance of an algorithm, all the pairs are mixed together without specifying separate covariate labels. In the following sections, we first present our experimental results on the overall protocol where covariate labels are not involved and then delve into the details of each covariate result.

\begin{figure*}
    \begin{center}
    \subfigure[ROC curves for IJB-B 1:1 covariates]{
            \includegraphics[width=0.47\textwidth]{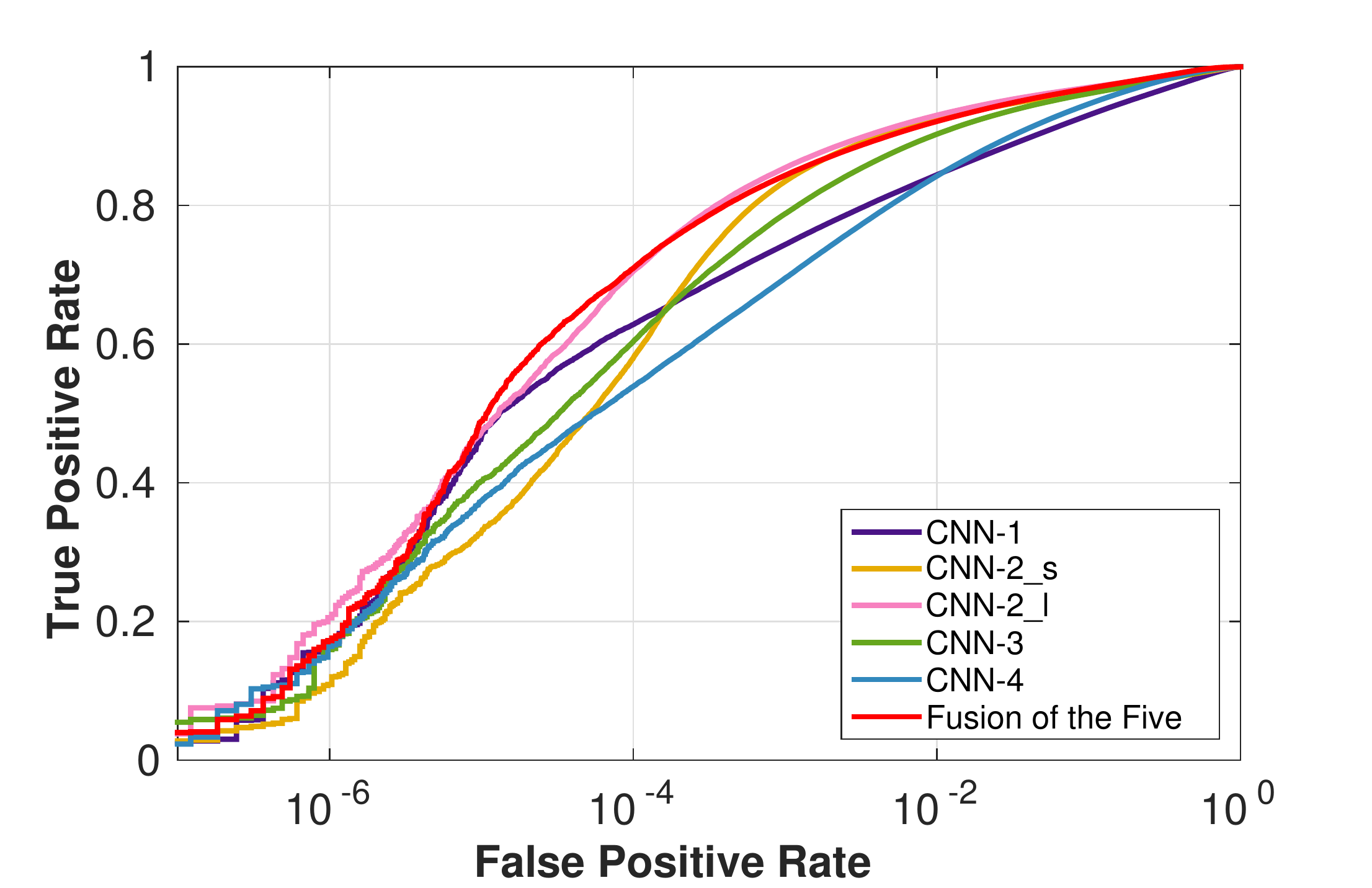}
            \label{ijbb_overall}
    }
    \subfigure[ROC curves for IJB-C 1:1 covariates]{
            \includegraphics[width=0.47\textwidth]{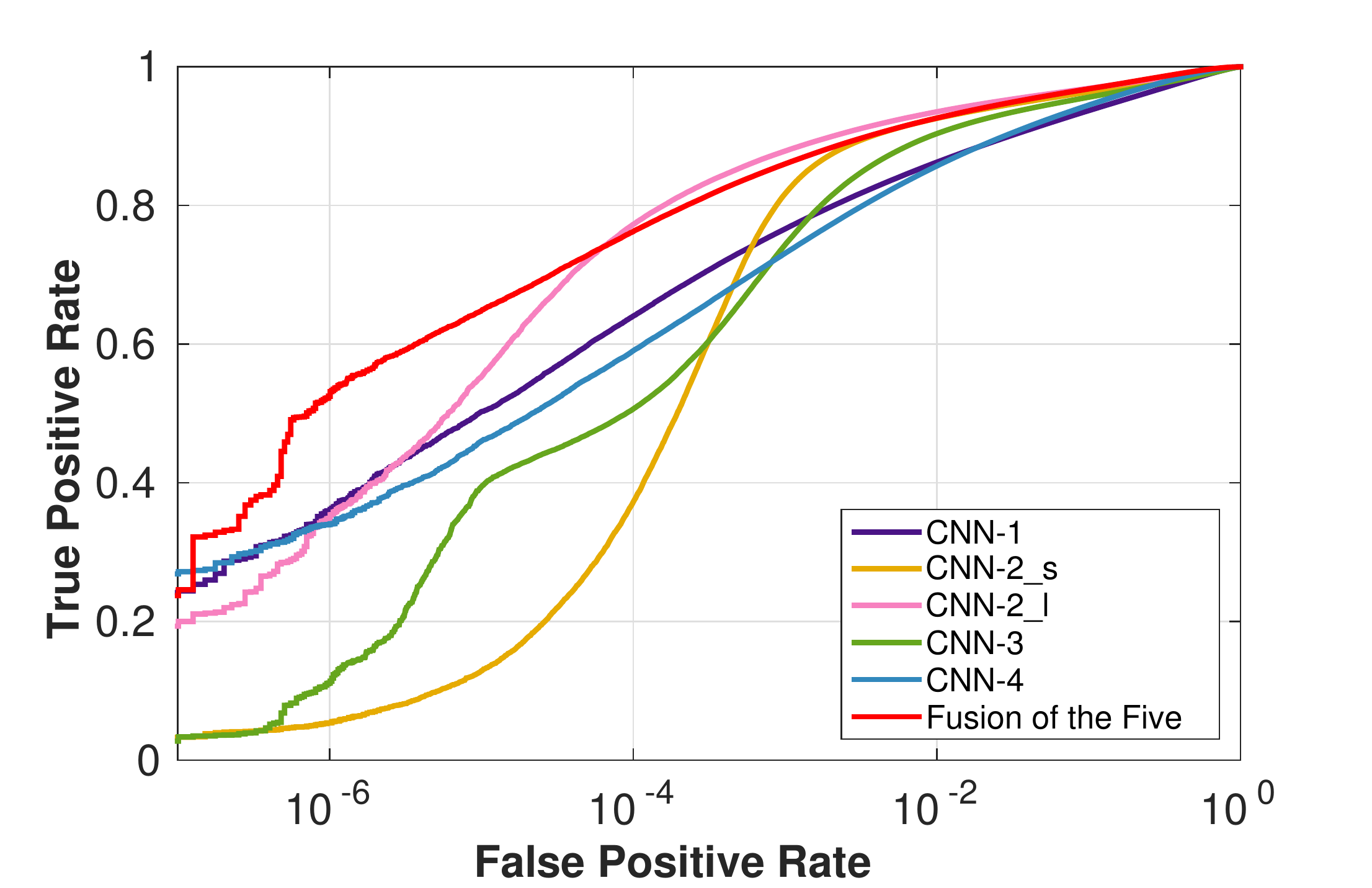}
            \label{ijbc_overall}
    }
    \caption{ROC curves for IJB-B and IJB-C 1:1 covariates overall protocol without specifying separate covariate labels. The fusion results are obtained by detection-score based fusion of the five CNN networks. The figures are best viewed in color.}
    \label{fig_overall}
    \end{center}
\end{figure*}

\begin{table*}[htbp]
\begin{center}
\resizebox{0.95\textwidth}{!}{
\begin{tabular}{|c|c|c|c|c|c|c|c|}
\hline
Method         & TAR@FAR = $10^{-7}$ & TAR@FAR = $10^{-6}$ & TAR@FAR = $10^{-5}$ & TAR@FAR = $10^{-4}$ & TAR@FAR = $10^{-3}$ & TAR@FAR = $10^{-2}$ & TAR@FAR = $10^{-1}$ \\ \hline
IJB-B before curation        & \textbf{0.0252}   & 0.1602     & 0.4455 & 0.6282 &  0.7474  & \textbf{0.8493}        & \textbf{0.9328}       \\ \hline
IJB-B after curation         & 0.0245  & \textbf{0.1731}        & \textbf{0.4636}   & \textbf{0.6284} & \textbf{0.7481} & 0.8447   & 0.9290          \\ \hhline{|=|=|=|=|=|=|=|=|}
IJB-C before curation        & 0.2417  & 0.3596     & 0.5023 & 0.6403 & 0.7660  & \textbf{0.8624}        & \textbf{0.9368}        \\ \hline
IJB-C after curation         & \textbf{0.2661}  & \textbf{0.3946}        & \textbf{0.5378}   & \textbf{0.6586} & \textbf{0.7684} & 0.8586   & 0.9337        \\ \hline
\end{tabular}
}
\end{center}
\caption{Performance comparison between before and after gender based training set curation on IJB-B and IJB-C 1:1 covariate overall protocol. All the results are generated using the CNN-1 architecture.}
\label{table_gender_clean}
\end{table*}

\begin{table*}[htbp]
\begin{center}
\resizebox{0.95\textwidth}{!}{
\begin{tabular}{|c|c|c|c|c|c|c|c|}
\hline
Method          & TAR@FAR = $10^{-7}$ & TAR@FAR = $10^{-6}$ & TAR@FAR = $10^{-5}$ & TAR@FAR = $10^{-4}$ & TAR@FAR = $10^{-3}$ & TAR@FAR = $10^{-2}$ & TAR@FAR = $10^{-1}$ \\ \hline
VGG-Face         & 0.0150 & 0.0440     & 0.0994 & 0.1515 &  0.2190  & 0.3318        & 0.5723        \\ \hline
Center-Face          & 0.0063 & 0.0353        & 0.0780   & 0.1363 & 0.2370 &   0.4206   & 0.7501        \\ \hline
Center-Face(retrain)          & 0.0517 & 0.1656       & 0.3880   & 0.6014 & 0.7620 &   0.8692   & 0.9460        \\ \hhline{|=|=|=|=|=|=|=|=|}
Fusion of our five model   & \textbf{0.0396} & \textbf{0.1707}   & \textbf{0.4882}    & \textbf{0.7093} & \textbf{0.8434} & \textbf{0.9213}   & \textbf{0.9688}      \\ \hline
\end{tabular}
}
\end{center}
\caption{Performance comparison for different methods on IJB-B 1:1 covariate overall protocol. Our fusion results are generated by detection score-based fusion of the five deep models. VGG-Face and Center-Face results are derived by applying their pretrained model to extract features and following the IJB-B 1:1 covariate overall protocol. Center-Face(retrain) is retrained using the curated MS-Celeb-1M dataset and the Center-Face model.}
\label{table_general_ijbb}
\end{table*}

\begin{table*}[htbp]
\begin{center}
\resizebox{0.95\textwidth}{!}{
\begin{tabular}{|c|c|c|c|c|c|c|c|}
\hline
Method          & TAR@FAR = $10^{-7}$ & TAR@FAR = $10^{-6}$ & TAR@FAR = $10^{-5}$ & TAR@FAR = $10^{-4}$ & TAR@FAR = $10^{-3}$ & TAR@FAR = $10^{-2}$ & TAR@FAR = $10^{-1}$ \\ \hline
VGG-Face        & 0.0513  & 0.0792     & 0.1159 & 0.1616 & 0.2275  & 0.3396       & 0.5918        \\ \hline
Center-Face     & 0.0479 & 0.0652     & 0.1005 & 0.1629 & 0.2746  & 0.4739       & 0.7733        \\ \hline
Center-Face(retrain)     & 0.2417 & 0.3596     & 0.5023 & 0.6403 & 0.7660  & 0.8624       & 0.9368        \\ \hline
\hhline{|=|=|=|=|=|=|=|=|}
Fusion of our five model  & \textbf{0.2371} & \textbf{0.5249}   & \textbf{0.6478}    & \textbf{0.7623} & \textbf{0.8599} & \textbf{0.9261}   & \textbf{0.9681}       \\ \hline
\end{tabular}
}
\end{center}
\caption{Performance comparison for different methods on IJB-C 1:1 covariate overall protocol. Our fusion results are generated by detection score-based fusion of the five deep models. VGG-Face and Center-Face results are derived by applying their pretrained model to extract features and following the IJB-C 1:1 covariate overall protocol. Center-Face(retrain) is retrained using the curated MS-Celeb-1M dataset and the Center-Face model.}
\label{table_general_ijbc}
\end{table*}


\subsection{Evaluation on the overall protocol}

\subsubsection{Results for five deep networks and score-level fusion}
To compare the performance of five deep networks, we present the ROC curves for each network and their score-level fusion. For detection score-based fusion, threshold $thr$ is set to 0.75 and the reweight coefficient $\alpha$ is set to 0.8. We also did a sensitivity analysis on these two parameters in Section~\ref{sec:ablation}. Figures~\ref{ijbb_overall} and~\ref{ijbc_overall} show the performance for IJB-B and IJB-C 1:1 covariates respectively. From both figures, we observe that CNN-2\_S performs very well at high FARs of the ROC curve, but the performance drops rapidly at low FARs. In contrast, CNN-1, CNN-3 and CNN-4 have smoother curves and perform better at low FARs but worse at high FARs than CNN-2\_S. Meanwhile, CNN-2\_L shows very strong performance for all FARs and outperforms the other four networks in middle range of FARs (FAR=$10^{-4}$, $10^{-3}$). Moreover, the fusion results of the five networks outperform all individual models, especially at low FARs of the ROC curve for the IJB-C dataset. This demonstrates the complementary behavior of the different models and fusion can always yield some improvements over individual models. By comparing the ROC curves of IJB-B and IJB-C datasets, we can see similar trends when FARs are larger than $10^{-4}$ but performance for IJB-B drops faster at low FARs of the ROC curve. In addition, at low FARs, different algorithms perform very differently for IJB-C but similarly for IJB-B dataset. This may be due to the IJB-B dataset containing more hard negative pairs that cannot be handled by any of these algorithms.

\begin{figure*}
    \begin{center}
    \subfigure[ROC curves with yaw difference changes for IJB-B]{
            \includegraphics[width=0.47\textwidth]{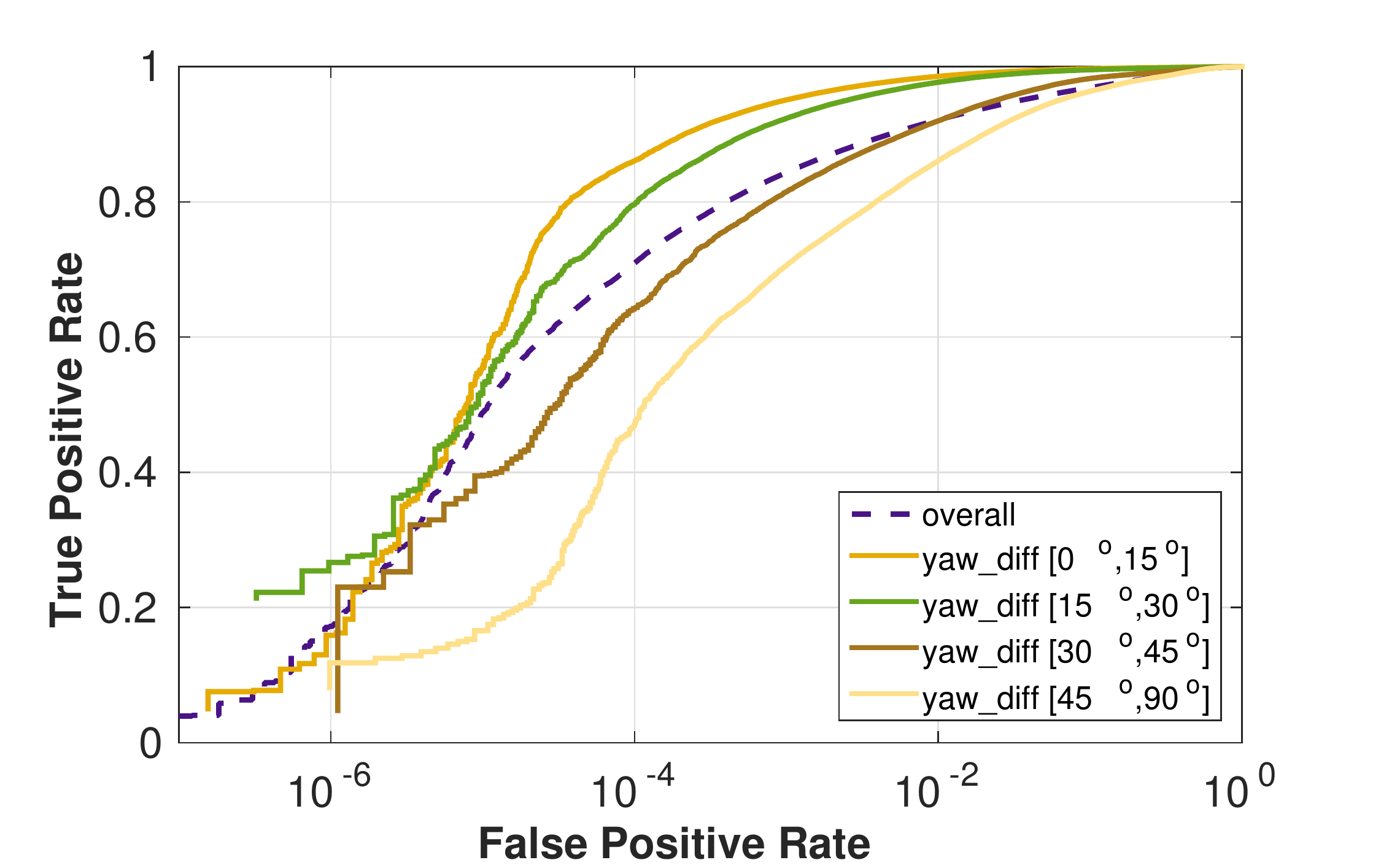}
            \label{fig_yaw}
    }
    \subfigure[ROC curves with absolute yaw angle changes for IJB-B]{
            \includegraphics[width=0.47\textwidth]{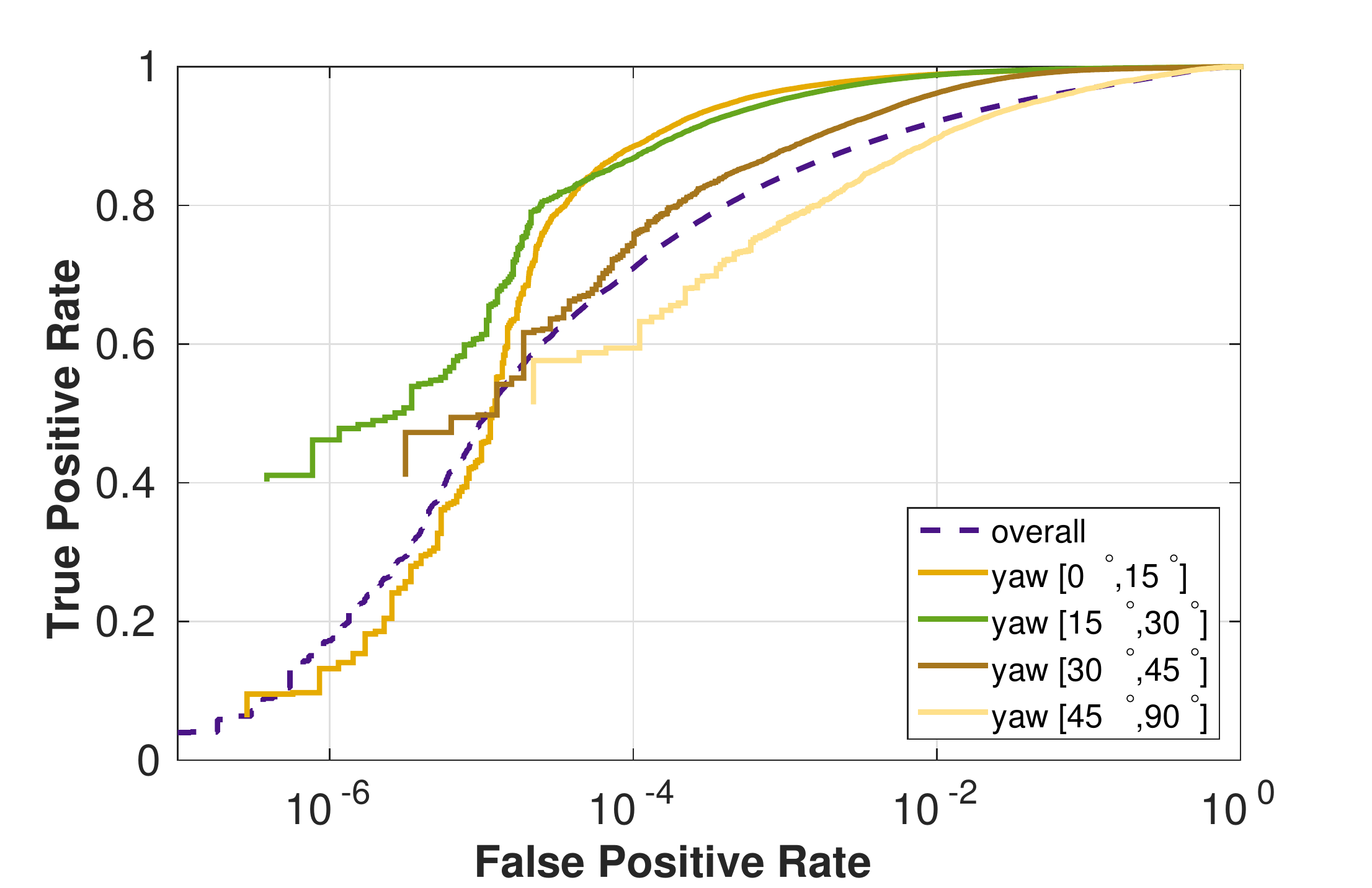}
            \label{yaw_absolute}
    }
    \caption{ROC curves when the yaw difference between two face images changes and when absolute yaw angle of faces changes. The range is from $0^\circ$ to $90^\circ$ because we average the features for original face and its mirrored image as the final face representation. The absolute yaw angles are computed by averaging two faces. The dashed line represents the results for the overall protocol.}
    \end{center}
\end{figure*}

\begin{figure*}
    \begin{center}
    \subfigure[ROC curves with roll difference changes for IJB-B]{
            \includegraphics[width=0.47\textwidth]{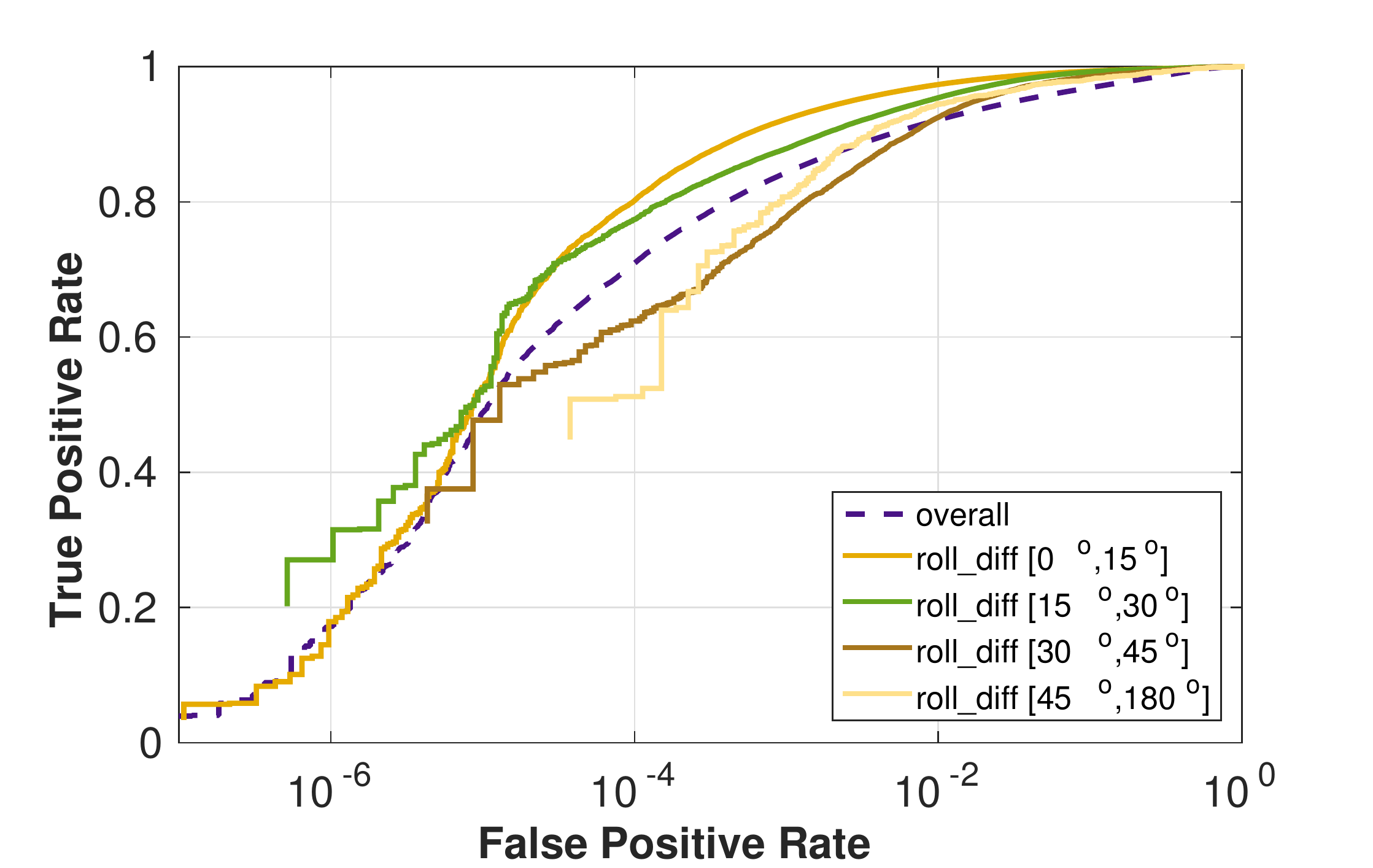}
            \label{fig_roll_ijbb}
    }
    \subfigure[ROC curves with roll difference changes for IJB-C]{
            \includegraphics[width=0.47\textwidth]{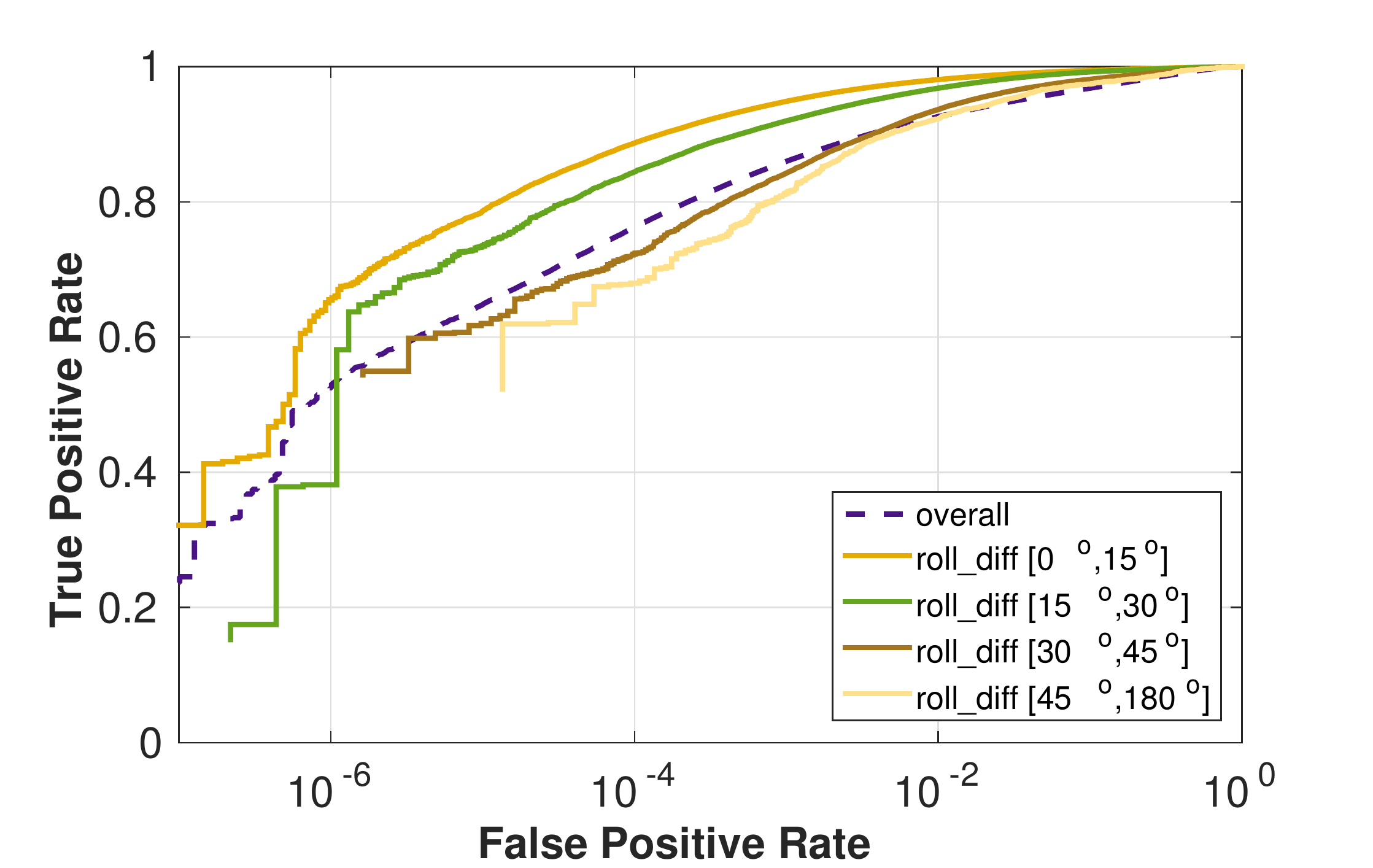}
            \label{fig_roll_ijbc}
    }
    \caption{ROC curves when the roll angle difference between two face images changes. The range is from $0^\circ$ to $180^\circ$. The dashed line represents the results for the overall protocol.}
    \label{fig_roll}
    \end{center}
\end{figure*}

\subsubsection{Performance improvement by gender based training set curation}
To test the efficiency of the dataset curation method discussed in Section~\ref{clean_gender}, we retrain CNN-1 using the training set curated by exploiting gender information and compare with results obtained before curation. From Table~\ref{table_gender_clean} it can be seen that the performance is improved at low FARs of ROC curves after training set curation on both IJB-B and IJB-C datasets. Since the goal of gender-based curation is to improve the model's capability to distinguish male and female subjects who looks very similar, performance improvements at low FARs are consistent with this goal because it indicates that the model can deal with hard negative pairs in a better way. On the other hand, we notice that the performance improvements on IJB-C are larger than on IJB-B, which means the gender information is more useful to detect the hard negative pairs in IJB-C than in IJB-B.

\subsubsection{Compared to other competitive methods}
We compare our fusion results with some other competitive methods and the performance for IJB-B and IJB-C are shown in Table~\ref{table_general_ijbb} and Table~\ref{table_general_ijbc} respectively. Although there exist many face networks (\eg, DeepID3~\cite{sun2015deepid3}, Pose-Aware Face Networks~\cite{masi2016pose}), their model are not publicly available. Therefore, we tested two widely used public models: VGG-Face~\cite{parkhi2015deep} and Center-Face~\cite{wen2016discriminative}. More specifically, we used the pretrained models provided by authors to extract features and followed their preprocessing steps on face images.  It is clearly shown in Table~\ref{table_general_ijbb} and Table~\ref{table_general_ijbc} that our fusion results outperform both VGG Face and Center-Face by large margins. There are two main reasons for this dramatic performance difference. First, we employ deeper models and various architectures to capture different characteristics of faces and conduct score-level fusion to further boost the performance. Second, the training set we use contains more faces with diverse face variations. In order to disentangle the effect of using different training sets, we retrain the Center-Face model using the curated MS-Celeb-1M dataset. The results for IJB-B and IJB-C dataset are shown in Table~\ref{table_general_ijbb} and Table~\ref{table_general_ijbc}. We can see significant improvements in performance compared to the pretrained model, but the proposed fusion method still outperforms the retrained model significantly. 
\begin{table}[htbp]
\begin{center}
\resizebox{0.95\linewidth}{!}{
\begin{tabular}{|c||c|c|c|c||c|c|c|c|c|c|}
\hline
             & \multicolumn{4}{c||}{Detection score threshold $thr$}    & \multicolumn{5}{c|}{reweighting ceofficient $\alpha$}    \\ \hline
TAR@FAR   & 0.70 & 0.75 & 0.80 & 0.85  & 0.75  & 0.80  & 0.85  & 0.90  & 0.95 \\ \hhline{|=||=|=|=|=||=|=|=|=|=|}
$10^{-7}$ &0.038 & 0.038 & 0.037 & 0.036  & 0.038 & 0.038 & 0.038 & 0.038 &0.038  \\ \hline
$10^{-6}$ &0.198 & 0.199 & 0.196 & 0.191  & 0.200 & 0.200 & 0.200 & 0.198 &0.186  \\ \hline
$10^{-5}$ &0.475 & 0.472 & 0.465 & 0.456  & 0.474 & 0.472 & 0.463 & 0.436 &0.410  \\ \hline
$10^{-4}$ &0.708 & 0.705 & 0.701 & 0.698  & 0.698 & 0.705 & 0.707 & 0.707 &0.705  \\ \hline
$10^{-3}$ &0.855 & 0.854 & 0.853 & 0.852  & 0.849 & 0.854 & 0.858 & 0.860 &0.861  \\ \hline
$10^{-2}$ &0.930 & 0.930 & 0.930 & 0.929  & 0.928 & 0.930 & 0.932 & 0.933 &0.934  \\ \hline
$10^{-1}$ &0.970 & 0.970 & 0.970 & 0.970  & 0.970 & 0.970 & 0.971 & 0.972 & 0.972  \\ \hline
\end{tabular}
}
\end{center}
\caption{Performance variations when detection score threshold $thr$ (left) and reweighting coefficient $\alpha$ (right) varies}
\label{table:thr_alpha}
\end{table}

\subsubsection{Parameter sensitivity analysis for detection-score based fusion}
\label{sec:ablation}

When we do the detection-score based fusion in~\ref{score_reweight}, there are two parameters: $thr$ and $\alpha$. Here, we present the results of an ablation study on the sensitivity of these two parameters and the performance for different parameter settings is shown in Table~\ref{table:thr_alpha}. All the performance is reported for IJB-B dataset using CNN-2. We observe that the threshold $thr$ does not severely affect the performance. In contrast, decreasing the reweighting coefficients $\alpha$ can significantly improve the performance at low FARs ($10^{-5}$, $10^{-4}$) while slightly decreasing the performance at high FARs. This supports the effectiveness of our fusion strategy.

\subsection{Evaluation on pose}
\label{sec_pose}
To evaluate the effects of pose variations on face verification performance, the protocol provides yaw and roll angles for each face. Since we use the average of the features for original face and its mirrored version as the final face representation, this restricts the range of yaw to $[0^\circ,90^\circ]$ and roll to $[0^\circ,180^\circ]$. Based on the yaw difference between a pair of faces, we divide all pairs into four groups: $[0^\circ,15^\circ]$, $[15^\circ,30^\circ]$, $[30^\circ,45^\circ]$, and $[45^\circ,90^\circ]$. Similarly, pairs are also divided into four groups based on roll difference: $[0^\circ,15^\circ]$, $[15^\circ,30^\circ]$, $[30^\circ,45^\circ]$, and $[45^\circ,180^\circ]$. 


From Figure~\ref{fig_yaw}, we observe that the yaw difference between a pair of faces significantly affect face verification performance. For both IJB-B and IJB-C, the ROC curves decrease monotonically as the yaw difference between two faces increases. Moreover, the performance drops much faster when the yaw difference is larger than $30^\circ$. This supports the following two findings: a) deep face representations are more robust to yaw changes (less than $30^\circ$) than traditional face representations such as LBP~\cite{ahonen2006face} (usually less than $15^\circ$); b) state-of-the-art deep networks are still sensitive to large yaw variations (larger than $30^\circ$). In addition to yaw difference between two faces, another key factor that may influence the performance is the absolute yaw value of faces. In other words, even if the yaw difference between two faces is relatively small (less than $15^\circ$), the performance may still be affected when the absolute yaw angles for both faces are large. In order to separate this factor from yaw difference, we further split the group of yaw difference $[0^\circ,15^\circ]$ into four subgroups based on their absolute yaw angles: $[0^\circ,15^\circ]$, $[15^\circ,30^\circ]$, $[30^\circ,45^\circ]$, and $[45^\circ,90^\circ]$, where the degrees are computed by averaging the absolute yaw angles of a pair of faces. The ROC curves are shown in Figure~\ref{yaw_absolute}. Similar to the effect of yaw difference, the absolute yaw angles of faces larger than $30^\circ$ causes a large performance drop while performance are not affected much when yaw is less than $30^\circ$. By comparing Figures~\ref{fig_yaw} and~\ref{yaw_absolute}, we have another interesting finding: performance for absolute yaw angles in $[45^\circ,90^\circ]$ and for yaw difference in $[45^\circ,90^\circ]$ are comparable, which means that as long as at least one of the two faces has extreme yaw angle, the performance will be poor. This result demonstrates that face images with extreme yaw angles ($[45^\circ,90^\circ]$) are hard for face matching regardless of the yaw difference because a large part of facial information is missing.

Figure~\ref{fig_roll} shows the face verification performance for various roll difference between two faces. We find that performance is better for groups whose roll differences are smaller than $30^\circ$. This result is surprising because in general the roll difference should not affect the face verification performance since 2D face alignment is performed before face matching to normalize all faces to have the same roll angle. However, the performance drop when increasing the roll difference shows that facial landmarks may not be accurate so that faces are not normalized as expected when the roll angle is large. 
\begin{figure*}
    \begin{center}
    \subfigure[ROC curves with different genders for IJB-B]{
            \includegraphics[width=0.47\textwidth]{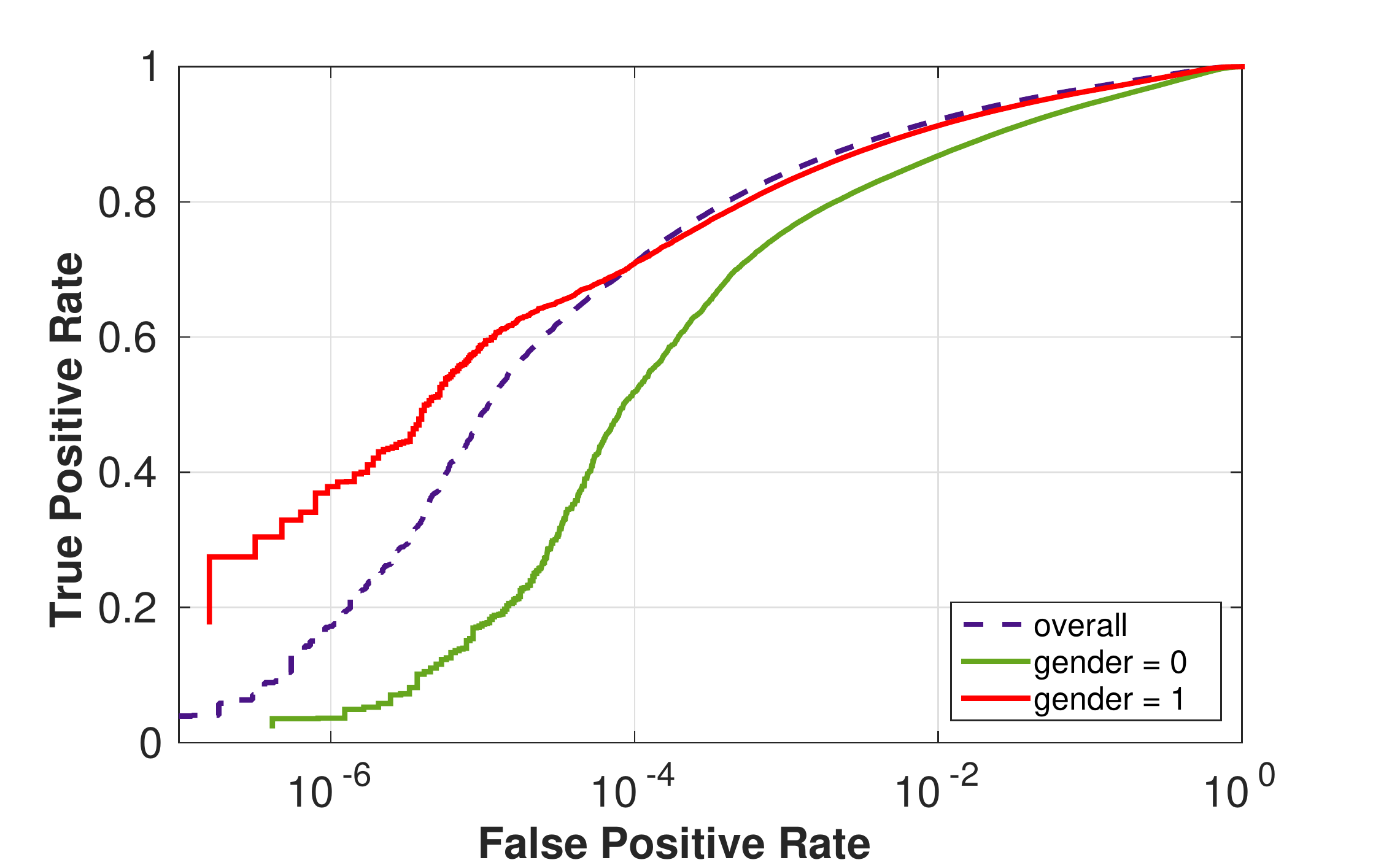}
            \label{fig_gender}
    }
    \subfigure[ROC curves with age changes for IJB-B]{
            \includegraphics[width=0.47\textwidth]{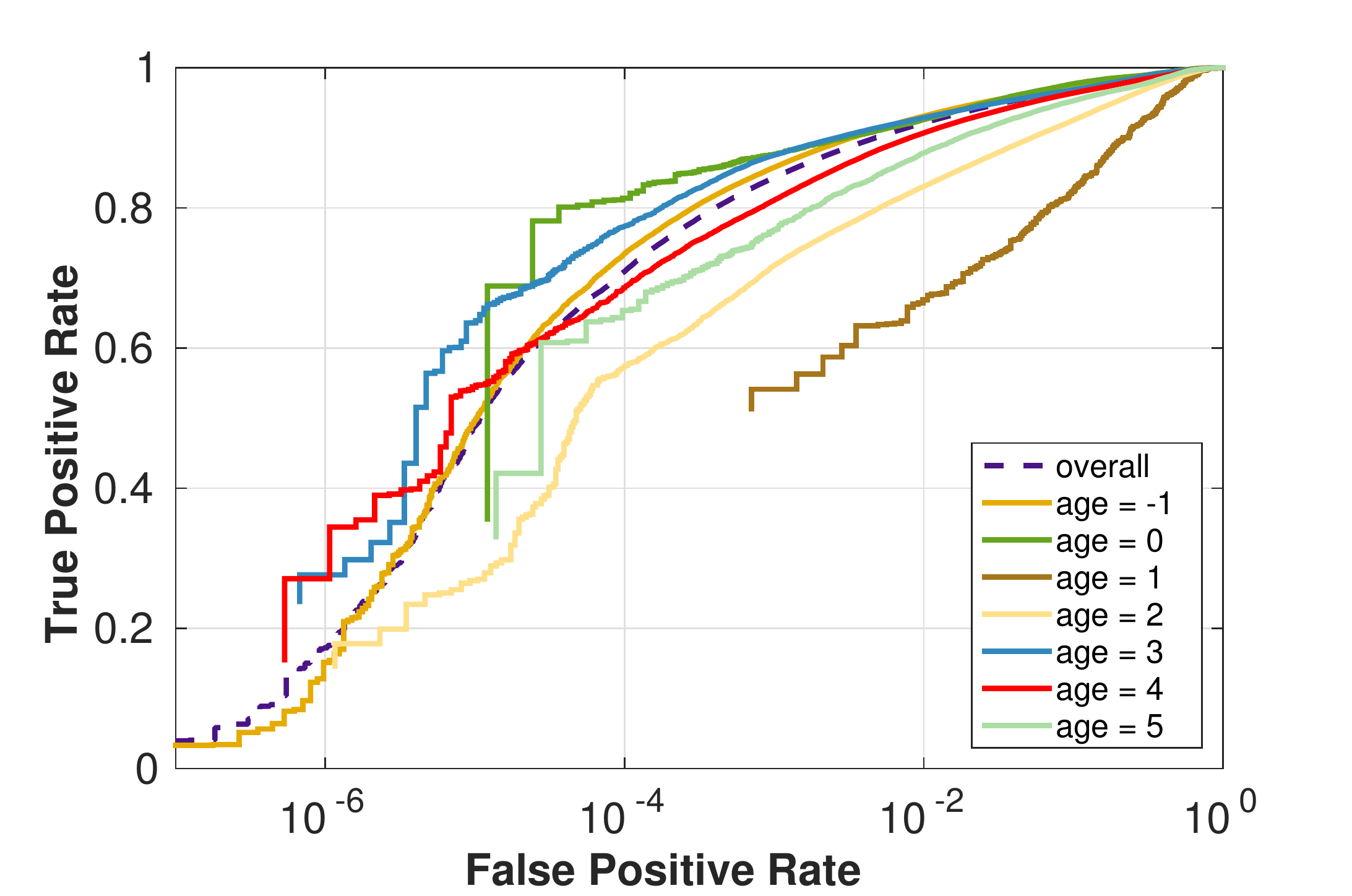}
            \label{fig_age}
    }
    \caption{ROC curves for different genders and for the case of age variation. For gender, female pairs are labeled as 0 and male pairs are labeled as 1. The dashed line represents the results for the overall protocol. For age covariate, ages in $[0,19]$, $[20,34]$, $[35,49]$, $[50,64]$, $65+$ are labeled as 1, 2, 3, 4, and 5 respectively. Ages that are different for two images in a pair are labeled as -1. Label 0 represents unknown ages for the pairs.}
    \end{center}
\end{figure*}


\subsection{Evaluation on gender}
\label{sec:gender}
In gender evaluation protocol, female pairs are assigned as label 0 and male pairs are assigned as label 1. In order to obtain valid ROC curves, the protocol does not consider the group when a pair of faces have different genders. This is because if two images are from different genders, they cannot form a positive pair and an ROC curve cannot be drawn without positive pairs. From Figure~\ref{fig_gender}, it can be observed that performance for males is much better than females' on both IJB-B and IJB-C datasets. A possible explanation for this result is that women's faces are often occluded by their long hair and their face appearance are changed by makeup.



\subsection{Evaluation on age}

The 1:1 covariate protocol labels the test pairs into seven categories based on their age distribution. Ages in $[0,19]$, $[20,34]$, $[35,49]$, $[50,64]$, $65+$ are labeled as 1, 2, 3, 4, and 5 respectively. Ages that are different for two faces in a pair are labeled as -1. Label 0 represents the group of unknown ages. Results for IJB-B dataset are shown in Figure~\ref{fig_age}. Due to space limitations, we did not include the IJB-C plots here because they show similar results as IJB-B. The dashed line represents performance for the overall protocol while the solid lines present curves for different age groups. It is shown that performance goes up when age groups change from 1 to 3. In contrast, the curves begin to fall from groups 3 to 5. It means the middle-age group (group 3) is the easiest one to be recognized while too young or too old subjects are both challenging for face verification. One possible explanation for this result may be because new born babies all look very similar and their unique facial features begin to emerge as they grow. However, when people become older, some common features for old people like wrinkles and sagging skins impair the uniqueness of their facial characteristics, which may make older subjects harder to be distinguished. In addition, we notice that age group -1 (ages of two images are different.) performs similarly as the the overall protocol, which means cross-age face verification is as hard as the general case. Nonetheless, this dataset does not fully explore the difficulty of cross-age face verification because the IJB-B and IJB-C datasets do not have images from the same person across large age gaps. 

\subsection{Evaluation on skin tone}

\begin{figure*}
    \begin{center}
    \subfigure[ROC curves with skin tone changes for IJB-B]{
            \includegraphics[width=0.47\textwidth]{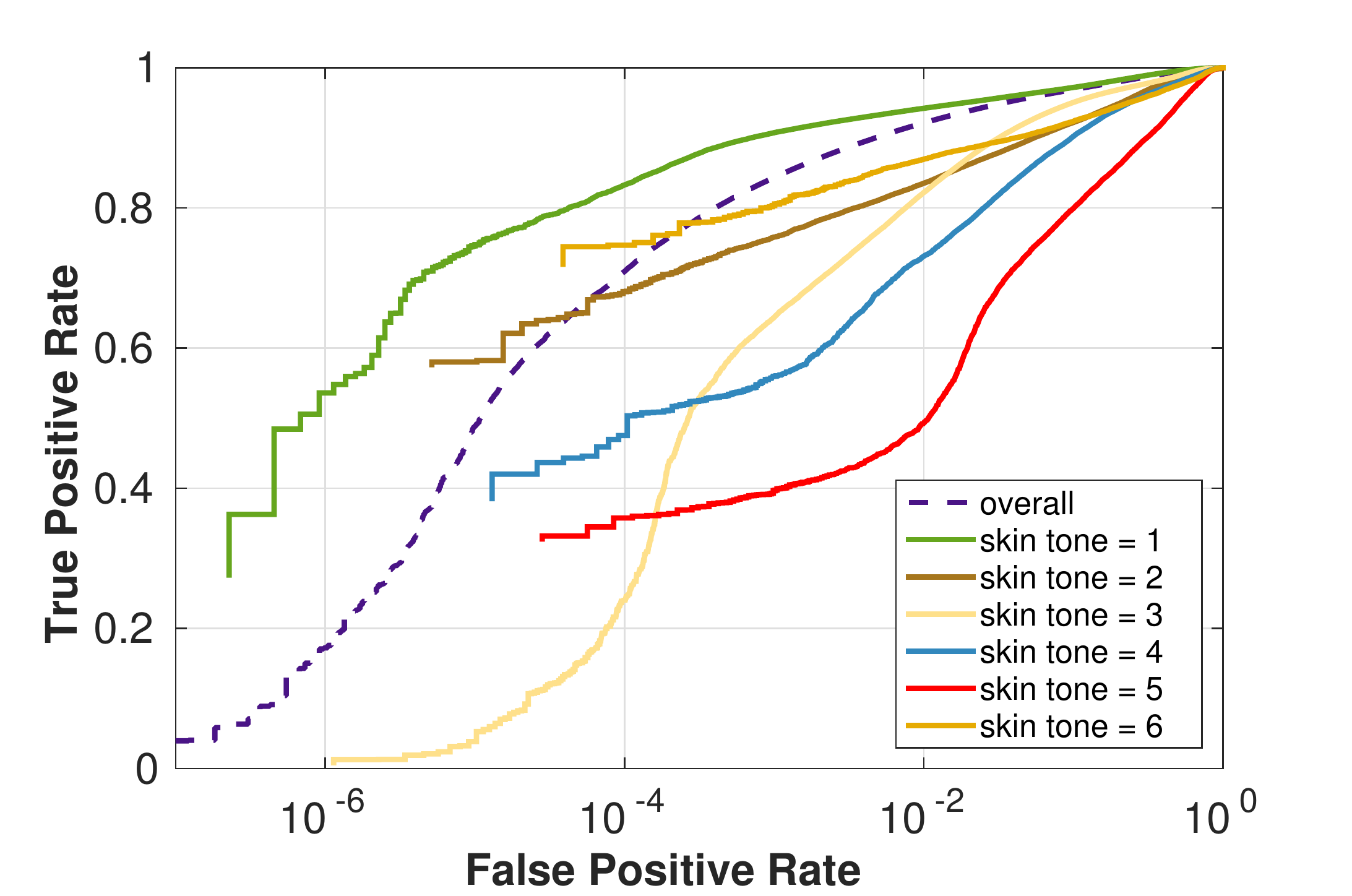}
            \label{fig_skintone_ijbb}
    }
    \subfigure[ROC curves with skin tone changes for IJB-C]{
            \includegraphics[width=0.47\textwidth]{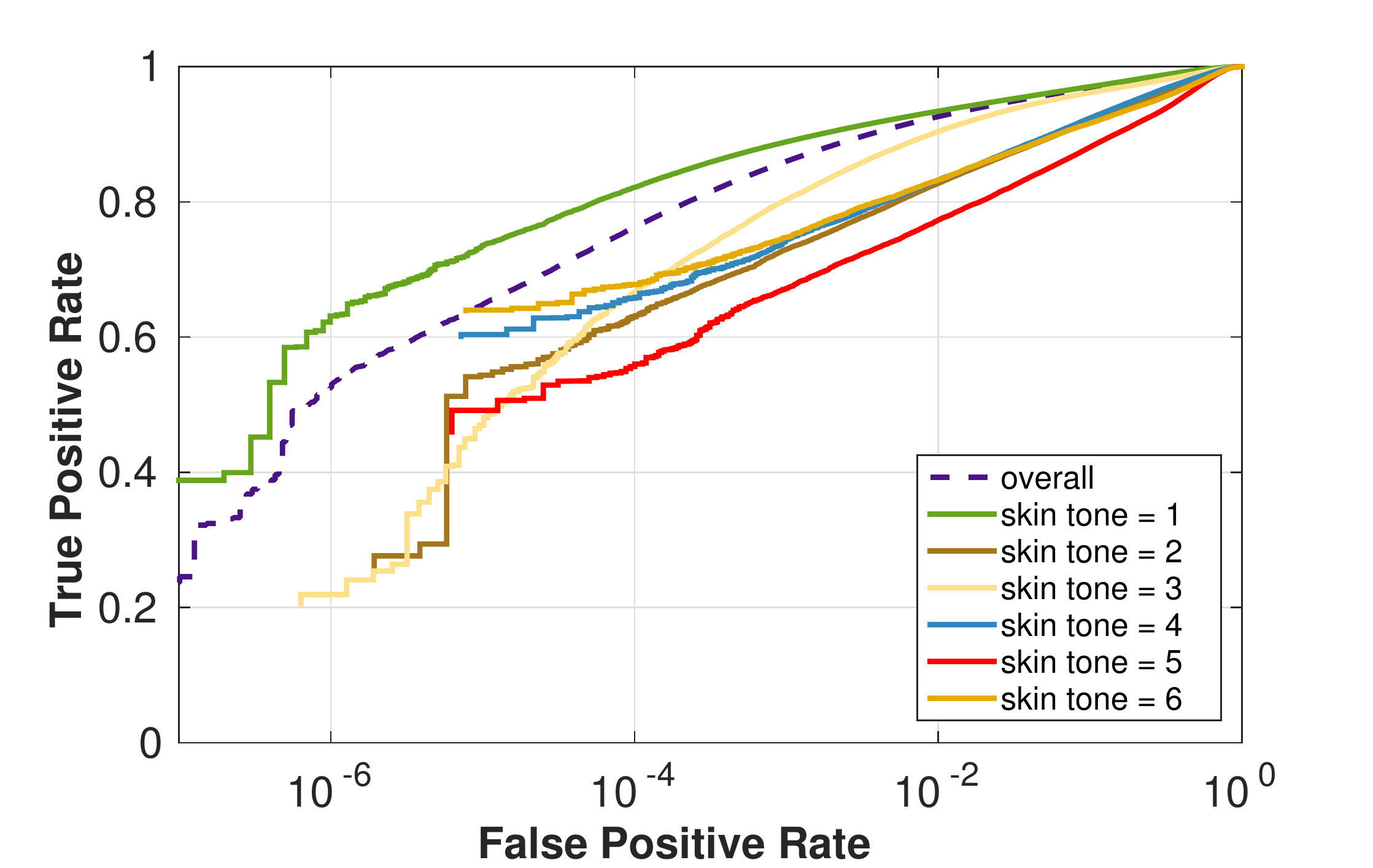}
            \label{fig_skintone_ijbc}
    }
    \caption{ROC curves with change in skin tone. The dashed line represents performance for the overall protocol while solid lines are curves for different skin tones. light pink, light yellow, medium pink/brown, medium yellow/brown, medium-dark brown and dark brown are labeled as 1, 2, 3, 4, 5, 6 respectively.}
    \label{fig_skintone}
    \end{center}
\end{figure*}

For skin tone, the protocol defines six classes: (1) light pink, (2) light yellow, (3) medium pink/brown, (4) medium yellow/brown, (5) medium dark brown, and (6) dark brown. Similar to gender, skin tone also does not contain group -1 because two images with different skin tones cannot form a positive pair. From Figure~\ref{fig_skintone}, we observe that performance for different skin tone groups show different trends on IJB-B and IJB-C. For IJB-B, the ROC curves for different groups are well separated. Since the skin tone change from light to dark for group 1 to 6, a general trend is that performance falls when the skin tone becomes darker. However, a counterexample is skin tone group 6 (darkest), which performs better than group 2 to group 5. In contrast, for IJB-C, except group 1 and group 5 which have the same trends as IJB-B, the performance for other skin tone groups is very close. Thus, we can only draw the conclusion that skin tone group 1 is the easiest and skin tone group 5 is the hardest for face verification. However, since defining or recognizing skin tones is ambiguous sometimes, it is hard to decide which skin tone is easier for face verification only from these results. 

\begin{figure*}
    \begin{center}
    \subfigure[ROC curves with nose/mouth visibility changes]{
            \includegraphics[width=0.47\textwidth]{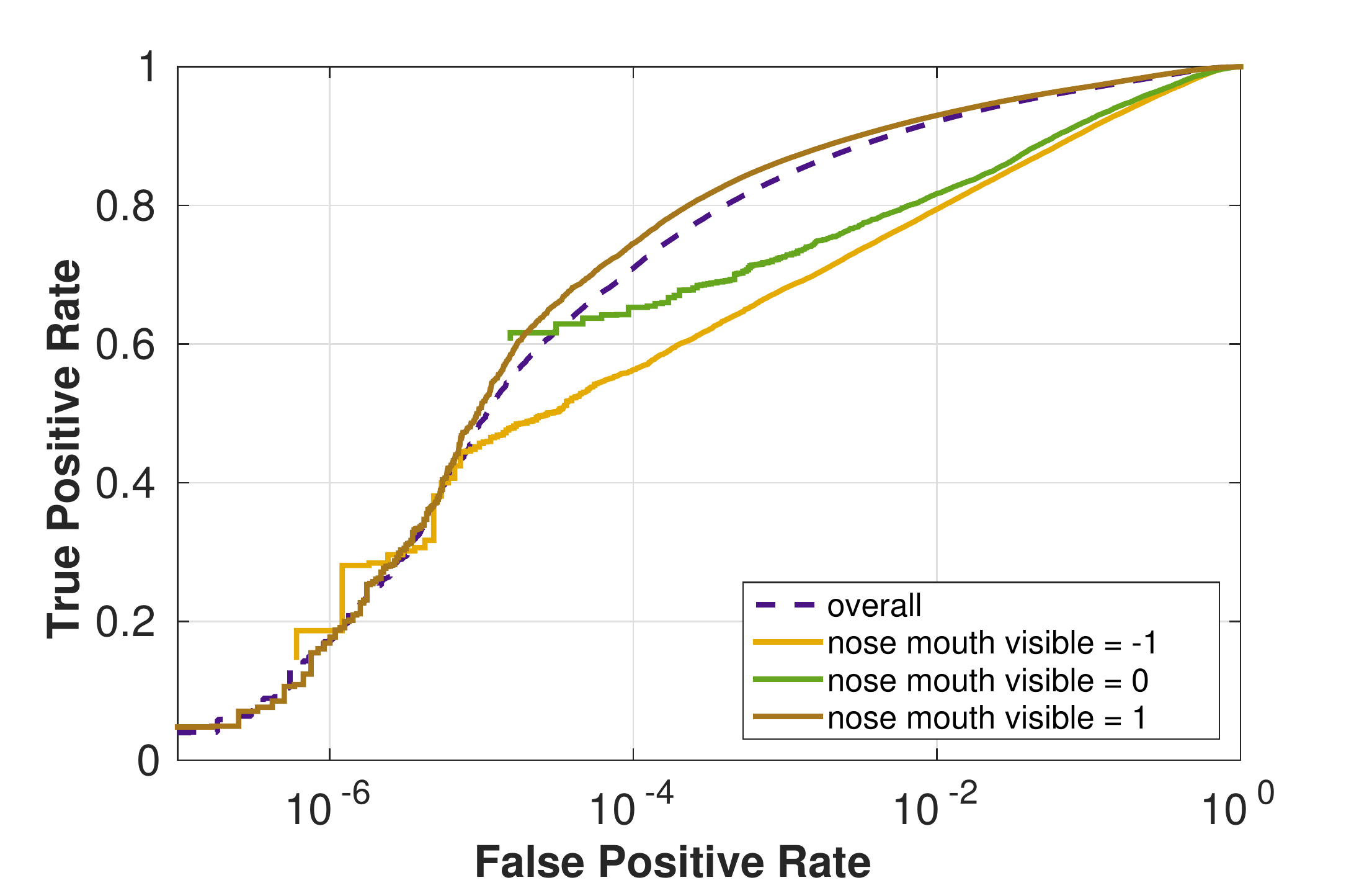}
            \label{fig_mouth}
    }
    \subfigure[ROC curves with forehead visibility changes]{
            \includegraphics[width=0.47\textwidth]{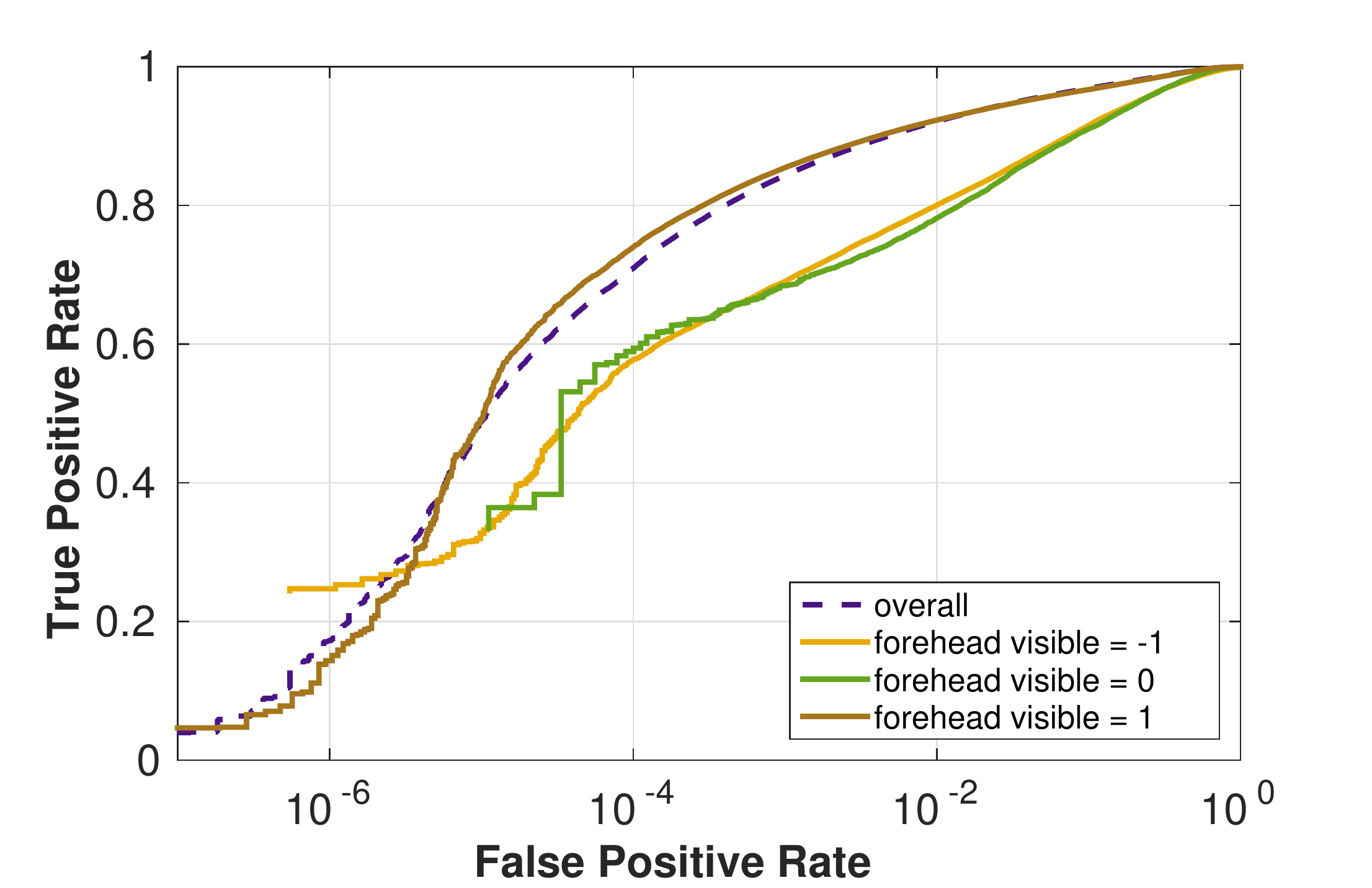}
            \label{fig_forehead}
    }
    \caption{ROC curves corresponding to nose/mouth and forehead visibilities for IJB-B dataset.}
    \label{fig_visibility}
    \end{center}
\end{figure*}

\begin{figure*}
    \begin{center}
    \subfigure[ROC curves with facial hair changes for IJB-B]{
            \includegraphics[width=0.47\textwidth]{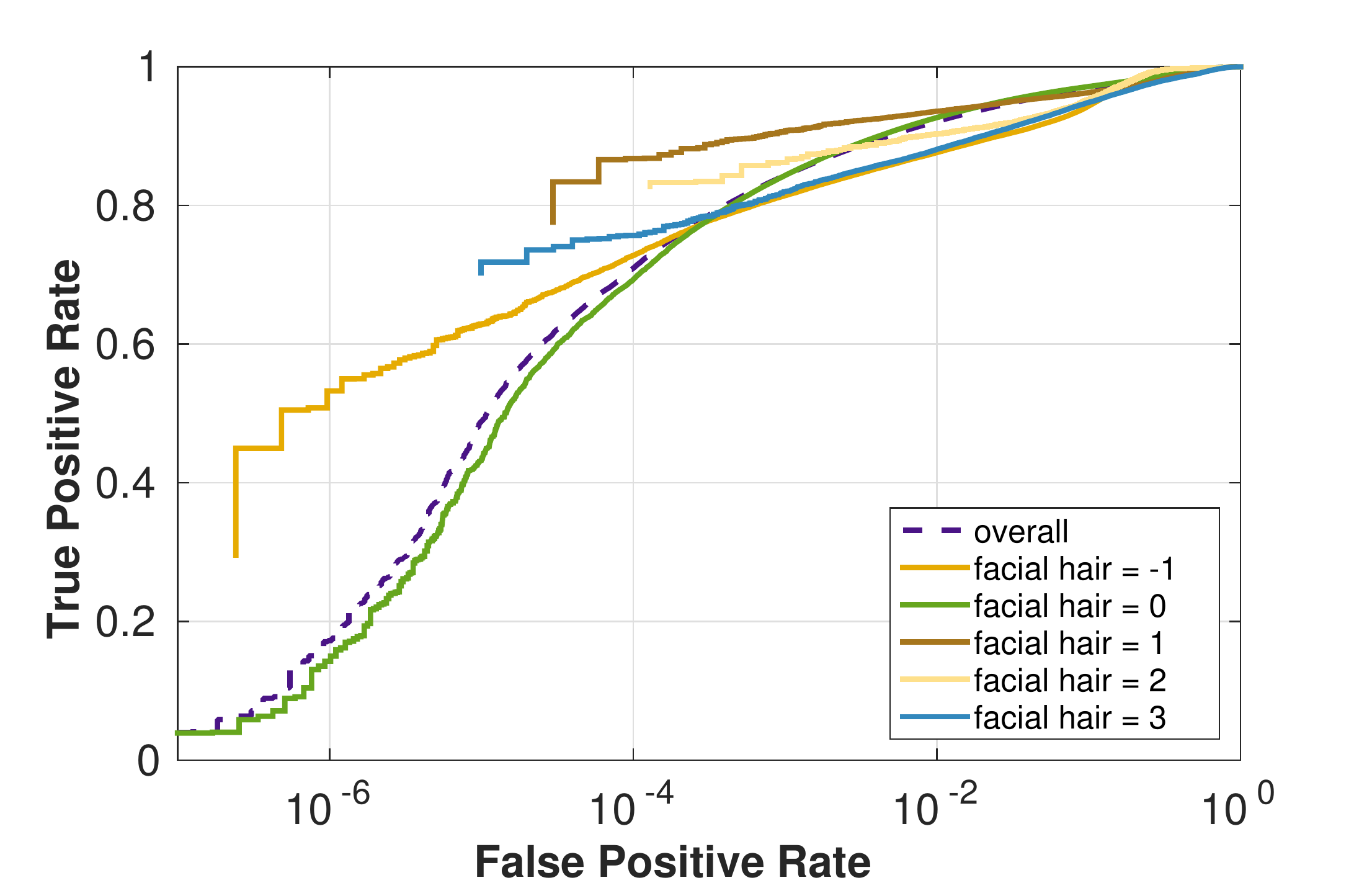}
            \label{fig_hair}
    }
   \subfigure[ROC curves with indoor/outdoor changes for IJB-B]{
            \includegraphics[width=0.47\textwidth]{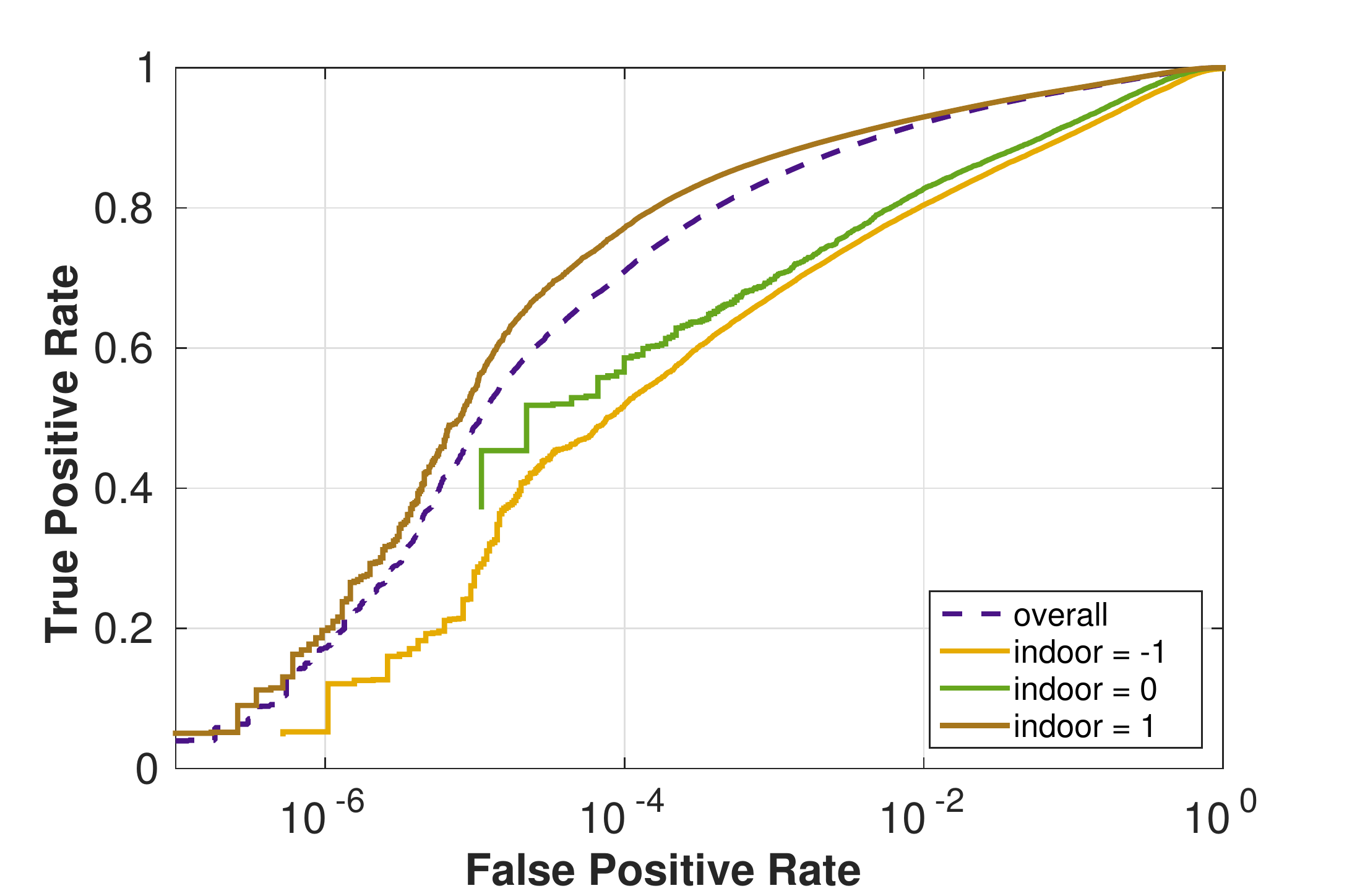}
            \label{fig_indoor}
    }
    \caption{ROC curves for the case of facial hair variation and for indoor/outdoor covariate. For facial hair, label 0 represents no facial hair, while label 1, 2, and 3 represent moustache, goatee and beard respectively. For indoor/outdoor, outdoor is labeled as 0 and indoor is 1. Label -1 means one image is taken indoor and the other outdoor.}
    \end{center}
\end{figure*}

\begin{figure}[htbp]
\begin{center}
 \includegraphics[width=3in]{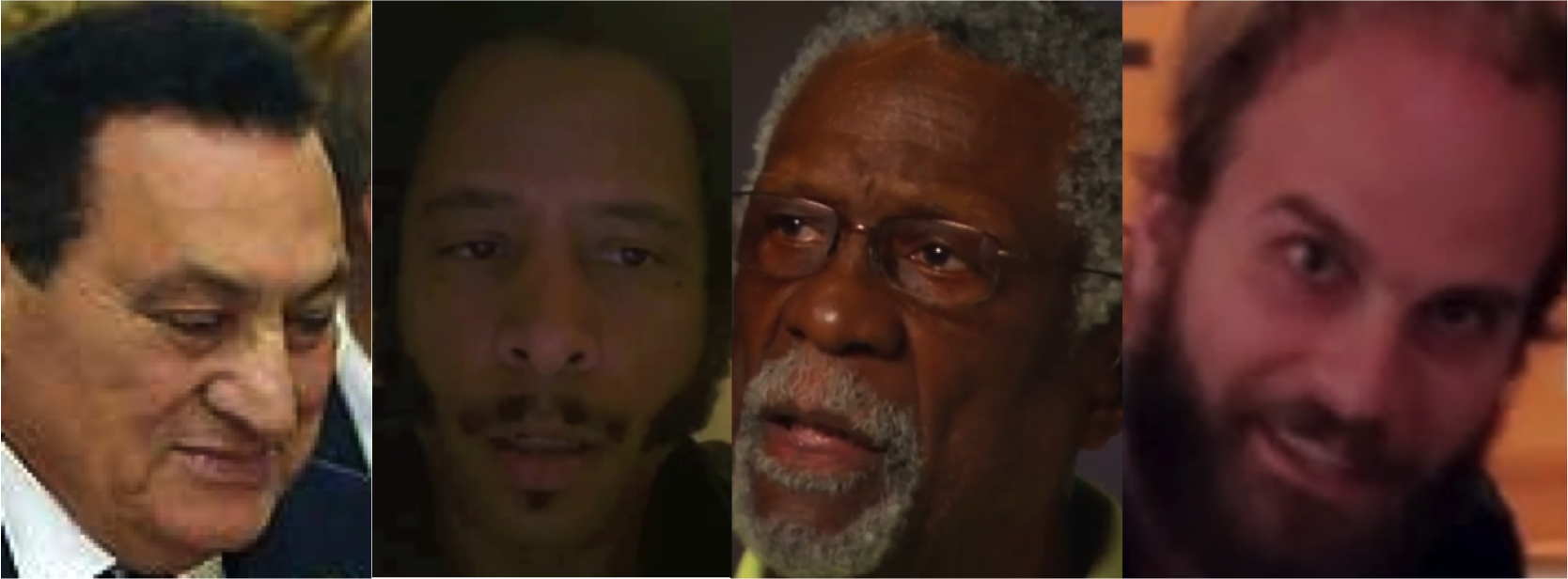}
 \caption{Sample images for different facial hair types. The four images corresponds to no facial hairs (0), moustache (1), goatee (2) and beard (3) respectively.}
 \label{fig_beard}
\end{center}
\end{figure}

\subsection{Evaluation on mouth and nose, and forehead visibility}

To evaluate the effects of occlusion, the protocol tests three types of visibilities for different facial parts: eyes visibility, mouth and nose visibility, and forehead visibility. Label 0 (1) represents the part is invisible (visible) for two images, and label -1 means the part is visible for one image but not for the other. The ROC curves for mouth and nose, and forehead visibility of IJB-B dataset are presented in Figures~\ref{fig_mouth} and~\ref{fig_forehead} respectively. Due to space limitations, we did not include the IJB-C plots here because they show similar results as IJB-B. In Figures~\ref{fig_mouth} and~\ref{fig_forehead}, similar results are shown for mouth/nose and forehead visibility: class -1 and 0 have comparable performance but are worse than class 1, which means that performance falls by large margins if nose, mouth or forehead are occluded for at least one of the images. This result indicates the importance of the visibility of key facial parts for recognizing faces.

\subsection{Evaluation on facial hair}

There are four classes for evaluation in facial hair protocol: class 0 represents no facial hair, while class 1, 2, and 3 represent moustache, goatee and beard respectively. Label -1 means facial hair classes are different for two images. Some sample images for moustache, goatee and beard are shown in Figure~\ref{fig_beard}. From Figure~\ref{fig_hair}, we observe that performance is not very sensitive to facial hair changes. This result demonstrates that facial hair does not change the key features of faces and state-of-the-art deep models can handle most facial hair variations.

\subsection{Evaluation on indoor/outdoor}

The last covariate we evaluate in the protocol is indoor/outdoor. Outdoor is labeled as 0 and indoor is 1. Label -1 means one image is taken indoor and the other outdoor. Performance is shown in Figure~\ref{fig_indoor}. We can see that the performance of class 1 is much better than class 0 and -1. This implies that indoor images are easier for face verification. There are two possible reasons for this result. First, outdoor images could be easily over-exposed and lose significant facial information. Second, outdoor images are often taken by hand-held cameras when people are walking. In contrast, indoor images are usually captured using tripod or at least without much motion. So the image quality for indoor images is often better than outdoor images.

\begin{figure*}
    \begin{center}
    \subfigure[ROC curves with age and gender changes on IJB-B]{
            \includegraphics[width=0.45\textwidth]{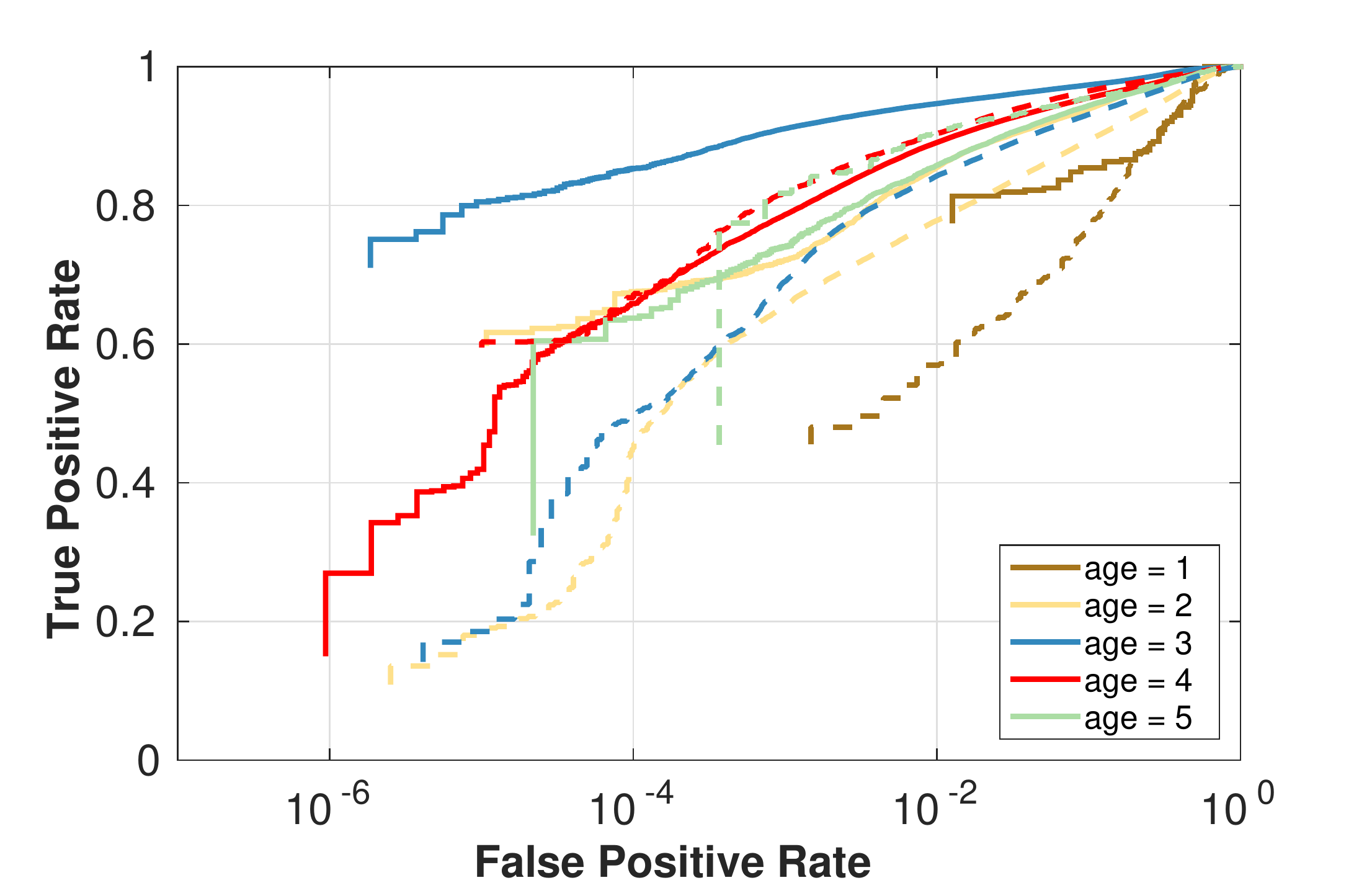}
            \label{fig_genders_age}
    }
    \subfigure[ROC curves with skin tone and gender changes on IJB-B]{
            \includegraphics[width=0.44\textwidth]{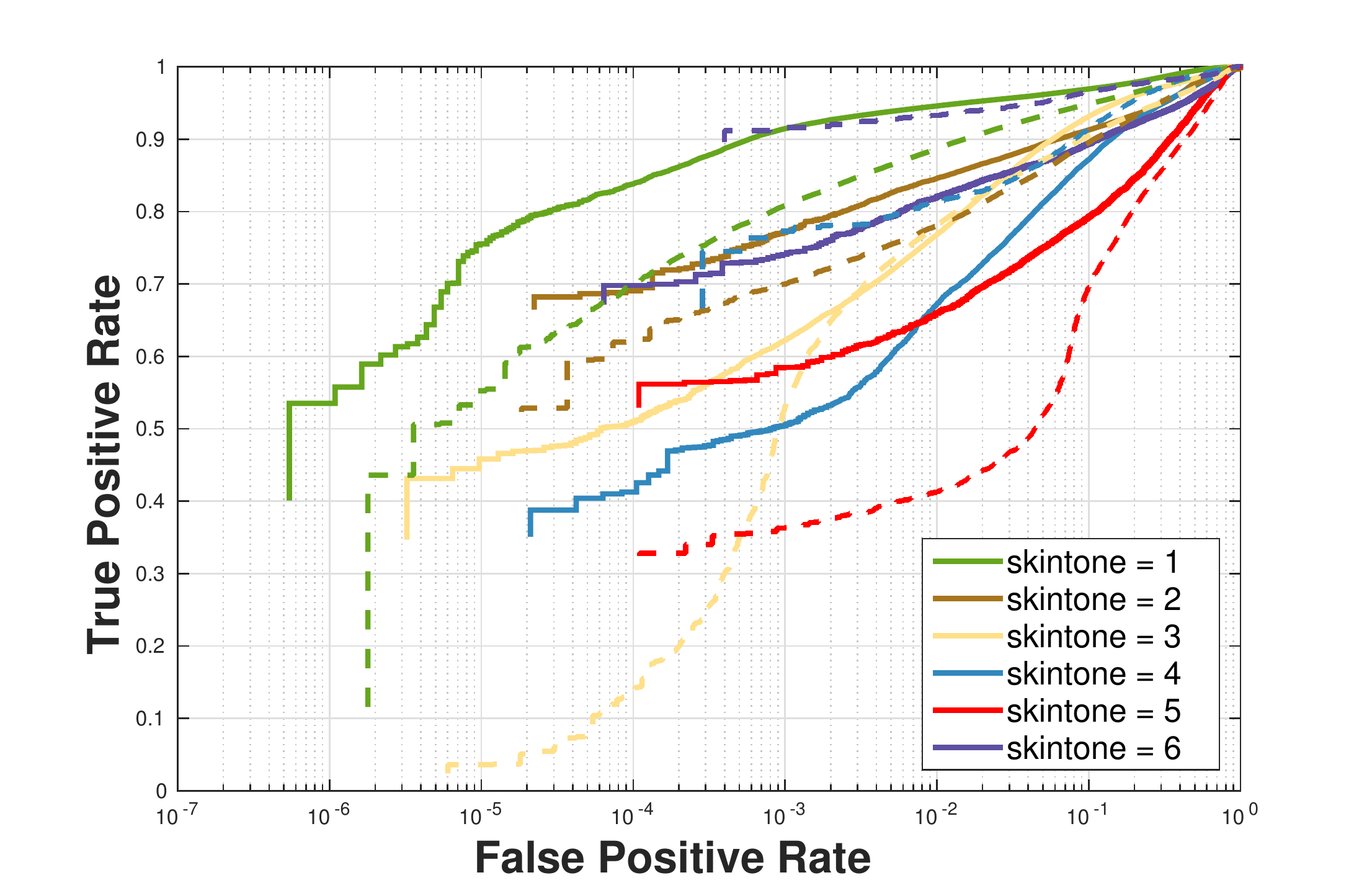}
            \label{fig_genders_skin}
    }
    \caption{ROC curves corresponding to age and gender (left) changes, and skin tone and gender (right) changes. Color lines represent different age groups and skin tones where small numbers represent young ages and light skin tones. Females is showed as dashed lines and solid lines represent males. }
    \label{fig_genders_age_skin}
    \end{center}
\end{figure*}

\begin{figure*}
    \begin{center}
    \subfigure[ROC curves with indoor/outdoor and nose and mouth visibility changes.]{
            \includegraphics[width=0.45\textwidth]{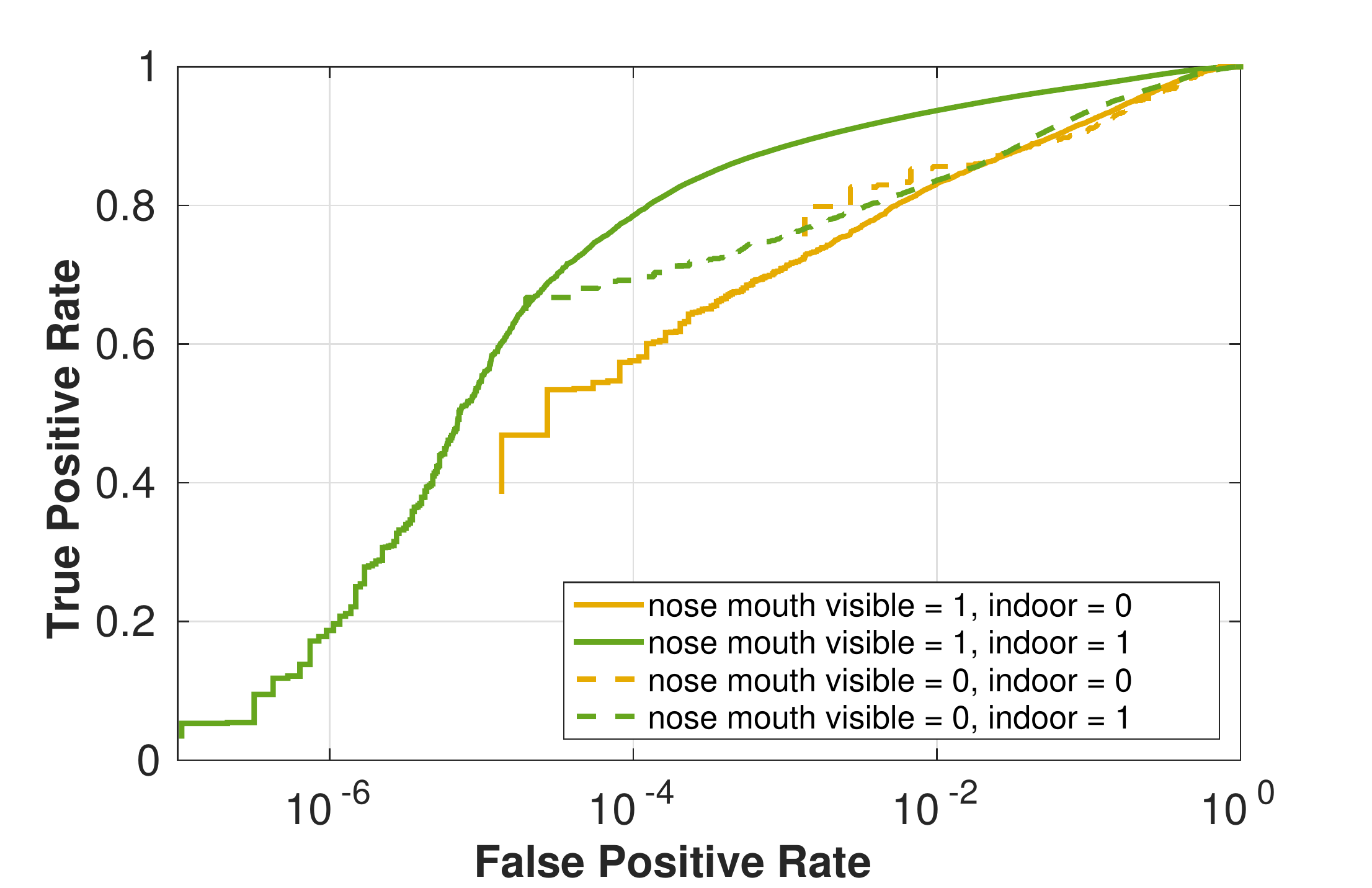}
            \label{fig_indoor_nose}
    }
    \subfigure[ROC curves with indoor/outdoor and forehead visibility changes.]{
            \includegraphics[width=0.45\textwidth]{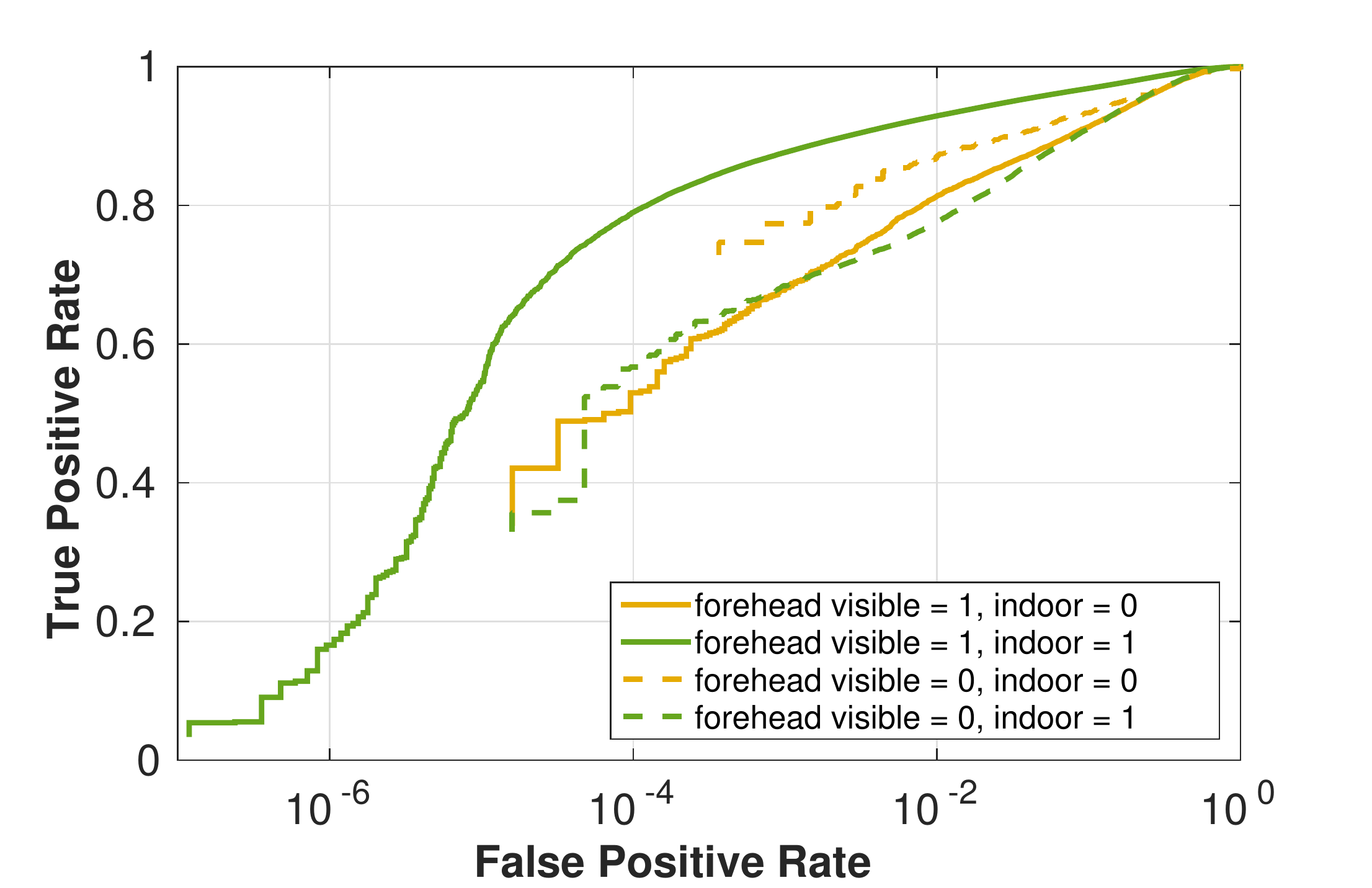}
            \label{fig_indoor_forehead}
    }
    \caption{ROC curves corresponding to visibility and indoor/outdoor on IJB-B.}
    \label{fig_indoor_vis}
    \end{center}
\end{figure*}

\begin{figure}[tb]
\begin{center}
 \includegraphics[width=3.2in]{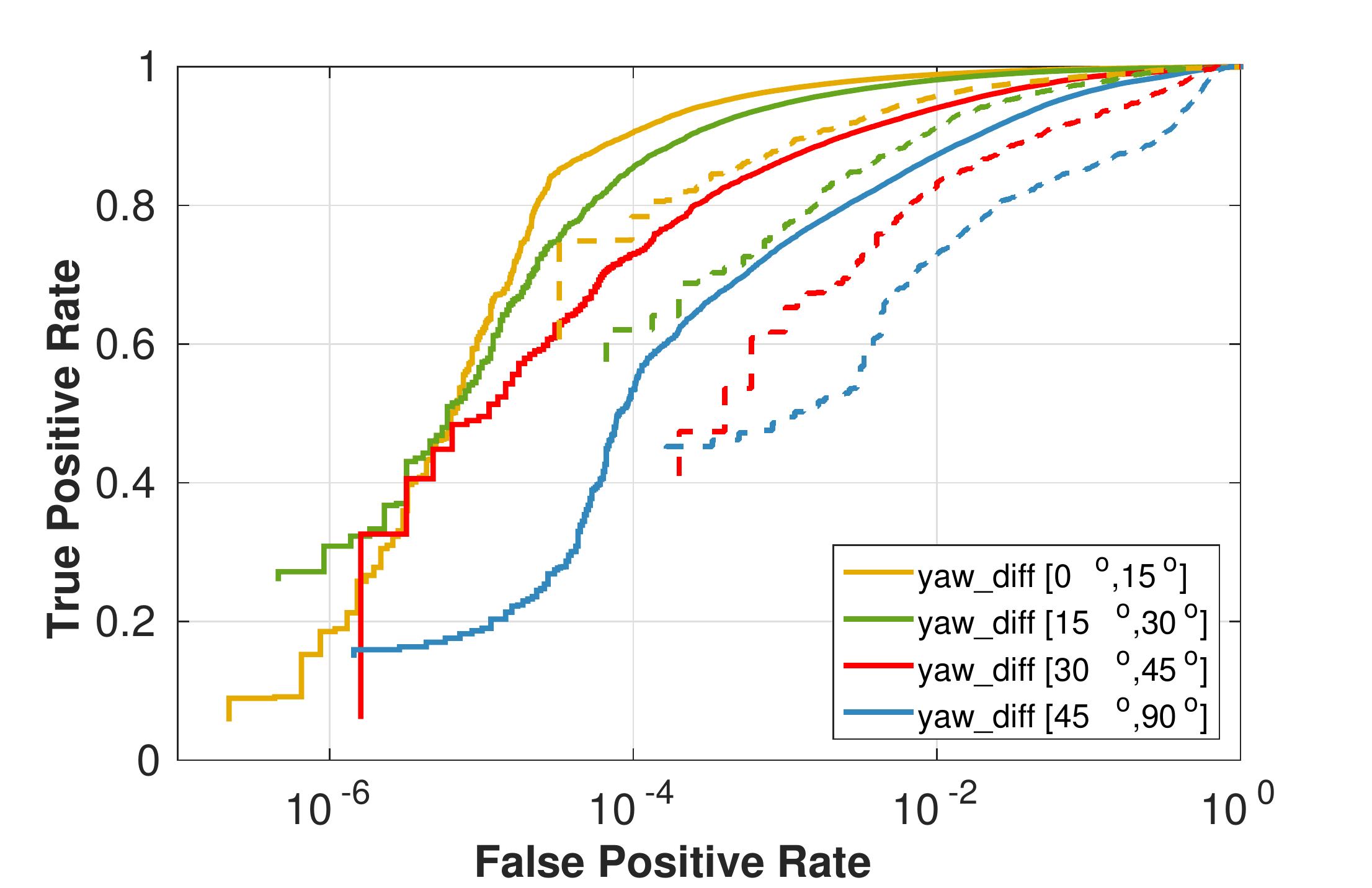}
 \caption{ROC curves corresponding to yaw difference and indoor/outdoor. Outdoor is showed as dashed lines and solid lines represent indoor.}
 \label{fig_indoor_yaw}
\end{center}
\end{figure}

\subsection{Evaluation on the effects of multiple covariates}

In the unconstrained environment, multiple face covariates are often correlated with each other which may affect the performance. It has been found that some covariates may show different trends on face verification performance when other covariates are considered together~\cite{phillips2003evaluation,beveridge2010frvt}. To study the correlation among the different covariates, we chose four pairs of related covariates and evaluated their interactive effects: gender and age, gender and skin tone, indoor (outdoor) and nose mouth/forehead visibility, indoor (outdoor) and yaw angle difference. Due to space limitation, all experimental results are reported for the IJB-B dataset.

\subsubsection{Evaluation on gender and age}

In order to show how gender and age influence each other, we draw the ROC curves in Figure~\ref{fig_genders_age} for each possible combination of values from genders and age groups. Different age groups are represented using different colors and male/female is showed in solid/dashed lines. First, we fix the gender factor and compare the performance of different age groups for males or females. We see that males and females show very different trends on age group effects. More specifically, for males middle age group $[35,49]$ performs best and the performance for old age group $[50,64]$ and $65+$ decreases. In contrast, for females the performance always increases when age groups get older. 

Alternatively, we can fix the age group factor and compare the performance of males and females for each age group. As observed in Section~\ref{sec:gender}, in general, males achieve superior performance than females. However, this finding does not hold for age group $[50,64]$ and $65+$. For age group $[50,64]$, males and females perform comparably while for age group $65+$ females outperform males.

\subsubsection{Evaluation on gender and skin tone}

We repeated the procedure discussed above for analyzing the combination of gender and skin tone. The ROC curves are shown in Figure~\ref{fig_genders_skin}. For skin tone groups 4 and 6, performance for females is better than that for males, while males performs better for group 1, 2 and 5. For skin tone group 3, males and females perform similar. This result shows that the combinations of gender and skin tone do not show clear trends and the performance is dependent on datasets.


\subsubsection{Evaluation on indoor (outdoor) and nose mouth / forehead visibility}

In addition to the demographic covariates, we are also interested in the mixed effects of covariates related to extrinsic factors. Figure~\ref{fig_indoor_vis} shows the performance for different indoor/outdoor and nose mouth/forehead visibility combinations. As we already saw, visible nose mouth/forehead and indoor are more favorable for better performance. However, these two factors may not have independent impacts on performance. From Figure~\ref{fig_indoor_vis}, we find that only when nose, mouth, or forehead is visible and the images are taken indoor, the performance is good. Either occlusion or outdoor can deteriorate the performance. 

\subsubsection{Evaluation on indoor (outdoor) and yaw angle difference}

The last combination we considered is indoor/outdoor and yaw angle difference. The ROC curves are presented in Figure~\ref{fig_indoor_yaw}. We notice that when fixing the indoor/outdoor factor, the performance for smaller yaw angle difference is always better. On the other hand, when the yaw angle difference is fixed, indoor faces always outperform outdoor faces. This result demonstrates that yaw angle difference and indoor/outdoor can affect the face verification performance independently and changing any one of the two factors can affect the performance.

\subsection{Evaluation on CFP dataset}

\begin{figure}[tb]
\begin{center}
 \includegraphics[width=3.2in]{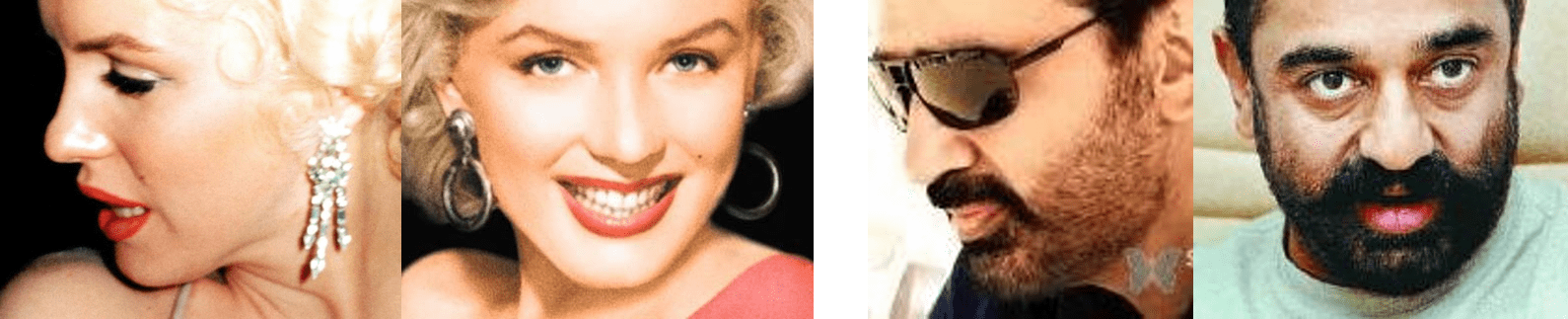}
 \caption{Sample images for CFP datasets.}
 \label{fig:sample_cfp}
\end{center}
\end{figure}

Since pose variation is a key challenging issue for face verification, we also used the Celebrities in Frontal-Profile (CFP) dataset to further investigate the underlying effects of extreme pose variations on unconstrained face verification performance. The CFP dataset consists of 7,000 still images from 500 subjects with 14 images per subject. For each subject, it has 10 images in frontal pose and 4 images in profile pose. To evaluate the performance for different poses, the protocol contains two settings: frontal-to-frontal (FF) and frontal-to-profile (FP) face verification. In the frontal-to-frontal settin{}g, two test images are both in frontal pose and in frontal-to-profile setting, a test pair includes one frontal face and one profile face. Each setting divides the whole dataset into ten splits and each split consists of 350 positive and 350 negative pairs. Some sample images are shown in Figure~\ref{fig:sample_cfp}.

\begin{table*}[htbp]
\begin{center}
\resizebox{0.95\textwidth}{!}{
\begin{tabular}{|c||c|c|c||c|c|c|}
\hline
             & \multicolumn{3}{c||}{Frontal-to-Frontal}    & \multicolumn{3}{c|}{Frontal-to-Profile}    \\ \hline
             & Accuracy     & EER          & AUC          & Accuracy     & EER          & AUC          \\ \hline
Deep features~\cite{sengupta2016frontal} &  0.964(0.007)  & 0.035(0.007) & 0.994(0.003) &  0.849(0.018) &  0.150(0.020) & 0.930(0.016) \\ \hline
Human~\cite{sengupta2016frontal}  &   0.962(0.007)  & 0.053(0.018)  &  0.982(0.011) & 0.946(0.011) & 0.050(0.011) & 0.989(0.005)   \\ \hline
CNN-1        & 0.988(0.002) & 0.012(0.004) & 0.999(0.001) & 0.938(0.012) & 0.062(0.013) & 0.986(0.005) \\ \hline
CNN-2\_S        & \textbf{0.997(0.003)} & \textbf{0.003(0.003)} & \textbf{1.000(0.000)} & \textbf{0.981(0.007)} & \textbf{0.018(0.007)} & \textbf{0.997(0.002)} \\ \hline
CNN-3        & 0.994(0.004) & 0.006(0.005) & \textbf{1.000(0.001)} & 0.969(0.009) & 0.029(0.011) & 0.994(0.003) \\ \hline
CNN-4        & 0.982(0.008) & 0.018(0.008) & 0.998(0.001) & 0.912(0.012) & 0.085(0.012) & 0.972(0.006) \\ \hline
Fusion       & 0.995(0.003) & 0.004(0.004) & \textbf{1.000(0.001)} & 0.973(0.006) & 0.027(0.008) & 0.996(0.002) \\ \hline
\end{tabular}
}
\end{center}
\caption{Performance comparison for different methods on CFP dataset. Our fusion results are generated by averaging the four deep models. }
\label{table_cfp}
\end{table*}

\subsubsection{Performance evaluation metrics}

We follow the performance evaluation metrics used in~\cite{sengupta2016frontal} and report three numbers for each setting:
Area under the curve (AUC), Equal Error Rate (EER) and Accuracy. AUC measures the area under ROC curves and ranges from 0 to 1 where higher value corresponds to better performance. EER is the point where the false accept rate is equal to false reject rate. It ranges from 0 to 1 with lower values indicating better performance. To get accuracy, we use an optimal threshold to classify all pairs and calculate the classification accuracy. For the optimal threshold, we chose the value that provides highest classification accuracy on the cross validation set.

\subsubsection{Results for frontal-to-frontal and frontal-to-profile protocols}

The experimental results for frontal-to-frontal and frontal-to-profile protocols are summarized in Table~\ref{table_cfp}.  CNN-1 to CNN-4 results are obtained by using the same models and same processing steps for IJB-B and IJB-C experiments. For the fusion part, since all detection scores for the images in CFP dataset is near 1, we simply average the similarity score for CNN-1 through CNN-4. Deep features and human results are directly cited from~\cite{sengupta2016frontal}. The performance is reported by averaging over ten splits. 

For the frontal-to-frontal setting, CNN-1 to CNN-4 all outperform both the deep features method and human performance in~\cite{sengupta2016frontal}. CNN-2\_S and CNN-3 perform similarly and their performance is slightly better than CNN-1 and CNN-4. Since performances of CNN-2\_S and CNN-3 have already saturated, fusion results for the five networks do not change much compared to CNN-2\_S or CNN-3. For the frontal-to-profile setting, different algorithms begin to show significant difference in performance. CNN-1 results are slightly worse than human performance but are 2\% better than CNN-4. On the other hand, CNN-2\_S and CNN-3 both surpass human performance by more than 2\%. Another interesting finding in the results is the performance comparison between frontal-to-frontal and frontal-to-profile settings. While performance for different algorithms do not vary too much in frontal-to-frontal protocol, the performance drops from frontal-to-frontal to frontal-to-profile is quite different among compared algorithms. Generally speaking, better algorithms are more robust to extreme yaw variations and always have smaller performance degradation for frontal-to-profile setting. In particular, CNN-2\_S has the smallest performance drop of 1.6\% from frontal-to-frontal to frontal-to-profile, which is similar to human performance. However, if we compare the results with Section~\ref{sec_pose}, even the best results are still severely affected by pose variations. This is because IJB-B and IJB-C datasets contain other challenging factors and pose variations can still degrade performance once combined with these factors. Therefore, even for state-of-the-art face models, there is still room to improve robustness to extreme pose variations. 


\section{Conclusion and Future Work}
\label{conclude}
In this paper, we conducted comprehensive experiments to study the effects of covariates on unconstrained face verification performance. We also curated the training data by exploiting gender information and achieved improved performance. Experimental results on the overall protocols of IJB-B and IJB-C covariate verification tasks show the outstanding performance of five implemented deep models and their score-level fusion. This demonstrates that these deep models are more robust to different variations of faces than previous methods. However, when we focus on each specific covariate, we found that many covariates still significantly affect the verification performance. Pose variations and occlusions are the top confounding factors that could cause performance drop by large margins. In addition, indoor performance is much better than outdoors. On the other hand, the difficulty of unconstrained face verification varies significantly for different demographic groups. Age, gender and skin tone all have shown impacts on performance. Specifically, males are easier to verify than females and old subjects generally performs better than young ones. For skin tone, light pink achieved best performance while medium-dark brown performs worst. However, since IJB-B and IJB-C show very different tendencies on skin tone groups, we may not be able to draw a clear conclusion on its effects.

Most of the findings discussed above confirm the findings of previous studies. However, there are also some new findings that were rarely mentioned by other studies or somewhat surprising. First, we found that verification performance does not increase monotonically as subjects get older. In contrast, performance begins to drop for age group of $[50,65]$ and $65+$. This result is different from most studies which claim older subjects are always easier to be recognized. However, since most of other studies did not have a sufficient number of older subjects to analyze, their results still make sense because middle age group performs better than children and teenagers. Second, we observed that extreme roll angle differences between faces still affect performance substantially. This result is unexpected as roll variations should be eliminated by face alignment. Therefore, we conclude that face alignment performance needs to get better when faces are in extreme roll angles. 

Finally, we investigated the mixed effects of multiple covariates. First, males and females show very different trends on the effects of age groups. For males, performance first increases then drops when age goes up while for females, older age groups always perform better. On the other hand, the interaction of gender and skin tone does not show clear trends. Second, when we consider indoor/outdoor and occlusion together, we found that indoor and nose mouth/forehead visibility must be satisfied simultaneously to achieve good performance. However, indoor/outdoor and yaw angle difference can affect the performance independently.

Some of the results from our studies show several promising research directions. First, apart from the yaw problem, we should also consider the influence of roll when designing face verification systems. This can be done by either improved face alignment or more robust feature extraction models. Second, since gender, age and skin tone all have significant impact on performance, we may collect the training set more carefully to improve the performance on certain demographic groups. Third, we show preliminary results on how to use gender estimation for training data curation. Other covariates like race may also be used in a similar way. Moreover, we may combine covariates with clustering method for improved curation performance.


%

\section*{Acknowledgment}

This research is based upon work supported by the Office of the
Director of National Intelligence (ODNI), Intelligence Advanced
Research Projects Activity (IARPA), via IARPA R\&D Contract No.
2014-14071600012. The views and conclusions contained herein are
those of the authors and should not be interpreted as necessarily
representing the official policies or endorsements, either expressed
or implied, of the ODNI, IARPA, or the U.S. Government. The U.S.
Government is authorized to reproduce and distribute reprints for
Governmental purposes notwithstanding any copyright annotation
thereon.

\ifCLASSOPTIONcaptionsoff
  \newpage
\fi



%
{\small
\bibliographystyle{ieee}
\bibliography{ref_boyu}
}

%

\begin{IEEEbiographynophoto}{Boyu Lu}
Biography text here.
\end{IEEEbiographynophoto}

\begin{IEEEbiographynophoto}{Jun-Cheng Chen}
Biography text here.
\end{IEEEbiographynophoto}


\begin{IEEEbiographynophoto}{Carlos D Castillo}
Biography text here.
\end{IEEEbiographynophoto}

\begin{IEEEbiographynophoto}{Rama Chellappa}
Biography text here.
\end{IEEEbiographynophoto}




\end{document}